\documentclass[10pt,twocolumn,letterpaper]{article}

\usepackage[pagenumbers]{iccv} %

\usepackage{xcolor}         %
\usepackage{colortbl}
\usepackage{graphicx}
\usepackage{amsmath,amsfonts,amssymb,amsthm}
\usepackage{color}
\usepackage{caption}
\usepackage{subcaption}
\usepackage{xspace}
\usepackage{array}
\usepackage{multirow}
\usepackage{wrapfig}
\usepackage{adjustbox}
\usepackage{makecell}   %
\usepackage{MnSymbol}

\definecolor{iccvblue}{rgb}{0.21,0.49,0.74}
\usepackage[pagebackref,breaklinks,colorlinks,allcolors=iccvblue]{hyperref}

\def\paperID{7307} %
\def\confName{ICCV}
\def\confYear{2025}

\title{Centaur: Robust End-to-End Autonomous Driving with Test-Time Training}

\author{Chonghao Sima$^{1,2\ast}$ \quad \quad \quad
Kashyap Chitta$^{3,4\ast}$ \quad \quad \quad
Zhiding Yu$^2$ \quad \quad \quad
Shiyi Lan$^2$\\
Ping Luo$^{1}$ \quad \quad
Andreas Geiger$^{3,4}$ \quad \quad
Hongyang Li$^1$ \quad \quad
Jose M. Alvarez$^2$\\
[2mm]
$^{1}$~The University of Hong Kong \quad
$^{2}$~NVIDIA \quad
$^3$~University of Tübingen \quad
$^4$~Tübingen AI Center
}

\newcommand{\bc}{\mathbf{c}}

\newcommand{\bp}{\mathbf{p}}

\newcommand{\bt}{\mathbf{t}}

\newcommand{\bx}{\mathbf{x}}

\newcommand{\figref}[1]{Fig.~\ref{#1}}
\newcommand{\secref}[1]{Section~\ref{#1}}
\newcommand{\eqnref}[1]{Eq.~\eqref{#1}}
\newcommand{\tabref}[1]{Table~\ref{#1}}

\makeatletter
\DeclareRobustCommand\onedot{\futurelet\@let@token\@onedot}
\def\@onedot{\ifx\@let@token.\else.\null\fi\xspace}
\def\eg{e.g\onedot} 
\def\ie{i.e\onedot}

\makeatother

\newcommand{\boldparagraph}[1]{\vspace{0.1cm}\noindent{\bf #1.}}

\definecolor{darkgreen}{rgb}{0,0.7,0}
\definecolor{darkyellow}{rgb}{0.8,0.8,0}
\definecolor{bittersweet}{rgb}{1.0, 0.44, 0.37}
\definecolor{amber}{rgb}{1.0, 0.49, 0.0}
\definecolor{lgray}{rgb}{0.83,0.83,0.83}

\definecolor{color_unlabled}{rgb}{0.0,0.0,0.0}
\definecolor{color_vehicle}{rgb}{0.0,0.0,0.56}
\definecolor{color_road}{rgb}{0.5,0.25,0.5}
\definecolor{color_redlight}{rgb}{1.0,0.0,0.0}
\definecolor{color_person}{rgb}{0.859,0.078,0.234}
\definecolor{color_roadline}{rgb}{0.613,0.914,0.195}
\definecolor{color_sidewalk}{rgb}{0.953,0.137,0.906}

\definecolor{ellisred}{rgb}{0.87,0.44,0.38} %
\definecolor{ellisgreen}{rgb}{0.69,0.90,0.52} %
\definecolor{elliscyan}{rgb}{0.29,0.77,0.74} %
\definecolor{ellisorange}{rgb}{0.89,0.55,0.28} %
\definecolor{ellisblue}{rgb}{0.41,0.61,0.86} %

\definecolor{Tab0}{HTML}{1F77B4}
\definecolor{Tab1}{HTML}{ff7f0e}
\definecolor{Tab2}{HTML}{2ca02c}
\definecolor{Tab3}{HTML}{d62728}
\definecolor{Tab4}{HTML}{9467bd}
\definecolor{Tab5}{HTML}{8c564b}
\definecolor{Tab6}{HTML}{e377c2}
\definecolor{Tab7}{HTML}{7f7f7f}
\definecolor{Tab8}{HTML}{bcbd22}
\definecolor{Tab9}{HTML}{17becf}

\definecolor{Tabx0}{HTML}{4e79a7}
\definecolor{Tabx1}{HTML}{f28e2b}
\definecolor{Tabx2}{HTML}{e15759}
\definecolor{Tabx3}{HTML}{76b7b2}
\definecolor{Tabx4}{HTML}{59a14f}
\definecolor{Tabx5}{HTML}{edc948}
\definecolor{Tabx6}{HTML}{b07aa1}
\definecolor{Tabx7}{HTML}{ff9da7}
\definecolor{Tabx8}{HTML}{9c755f}
\definecolor{Tabx9}{HTML}{bab0ac}

\newcommand{\algname}{Centaur\xspace}

\newcommand{\uncname}{Cluster Entropy\xspace}

\begin{document}
\maketitle

{\let\thefootnote \relax \footnote{$^*$Equal contribution. \\
Primary contact:
\texttt{chonghaosima@connect.hku.hk}}}

\begin{abstract}
How can we rely on an end-to-end autonomous vehicle's complex decision-making system during deployment? One common solution is to have a ``fallback layer'' that checks the planned trajectory for rule violations and replaces it with a pre-defined safe action if necessary. Another approach involves adjusting the planner's decisions to minimize a pre-defined ``cost function'' using additional system predictions such as road layouts and detected obstacles. However, these pre-programmed rules or cost functions cannot learn and improve with new training data, often resulting in overly conservative behaviors. In this work, we propose Centaur (\textbf{C}luster \textbf{En}tropy for \textbf{T}est-time tr\textbf{A}ining using \textbf{U}nce\textbf{r}tainty) which updates a planner's behavior via test-time training, without relying on hand-engineered rules or cost functions. Instead, we measure and minimize the uncertainty in the planner's decisions. For this, we develop a novel uncertainty measure, called Cluster Entropy, which is simple, interpretable, and compatible with state-of-the-art planning algorithms. Using data collected at prior test-time time-steps, we perform an update to the model's parameters using a gradient that minimizes the Cluster Entropy. With only this sole gradient update prior to inference, Centaur exhibits significant improvements, ranking first on the \texttt{navtest} leaderboard with notable gains in safety-critical metrics such as time to collision. To provide detailed insights on a per-scenario basis, we also introduce \texttt{navsafe}, a challenging new benchmark, which highlights previously undiscovered failure modes of driving models. %
\end{abstract}

\section{Introduction}
\label{sec:intro}

\begin{figure}[t!]
    \centering
    \includegraphics[width=\columnwidth]{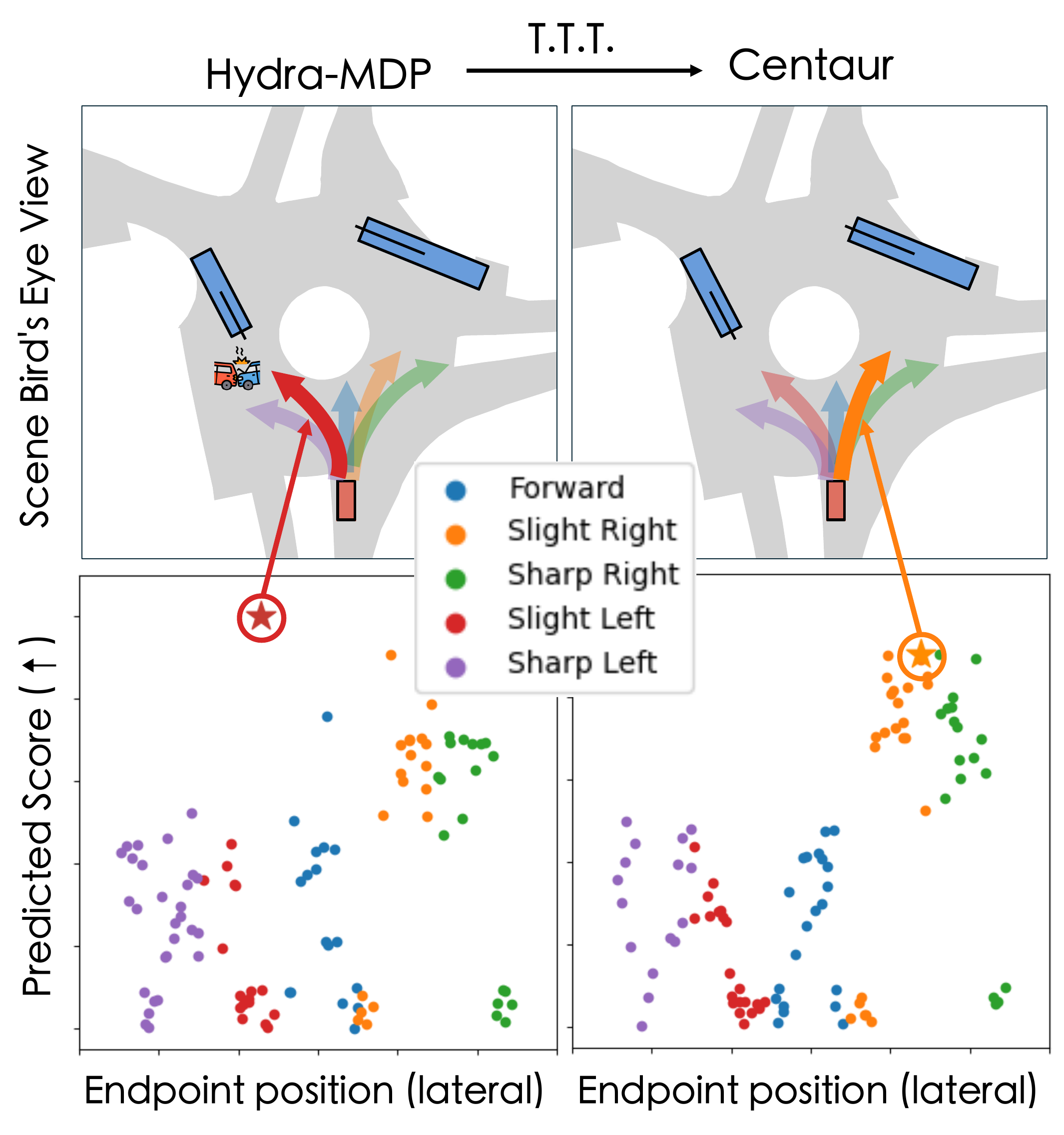}
    \caption{
    \textbf{Reducing entropy when uncertain with test-time training (TTT).} The end-to-end planner Hydra-MDP~\cite{li2024hydra}, state-of-the-art on \texttt{navtest}~\cite{dauner2024navsim}, predicts a score for every trajectory in a fixed set. In the scatter plots, we plot trajectory scores based on their lateral endpoint position, clustered into five categories (colored arrows in the upper visualizations). Hydra-MDP selects one trajectory with the highest predicted score ($\filledstar$) as its output. \textbf{Left:} In this roundabout, it selects a high-scoring outlier in the cluster `slight left' where the average score is low, indicating high uncertainty. \textbf{Right:} Such uncertainty is measured via our proposed \uncname, which we minimize via gradient descent (TTT) in Centaur. This leads to a more confident planner (with less entropy) that in this scene prefers `slight right' turns, avoiding a collision.
    }
    \label{fig:main-fig}
\end{figure}

As autonomous vehicles increasingly share public streets with human-driven vehicles, the need for safe interaction and robustness in unforeseen situations becomes vital~\cite{shalev2017safe,Liu2022CurseOR}. In particular, there is a rising interest in deploying end-to-end planners for autonomous vehicles (AVs)~\cite{chitta2023transfuser, hu2023uniad, Weng2024para, li2024hydra} due to their scalability and strong empirical performance in challenging scenarios~\cite{chen2023e2esurvey,li2023open}. However, in practice, end-to-end planners often apply fallback layers for real-world deployment, which use pre-defined safe actions to ensure safety in critical situations~\cite{chitta2023transfuser, vitelli2022safetynet, waywe2024ad2}. While useful, fallback layers require considerable domain expertise and manual engineering for their design, do not benefit from data-driven improvements, and constrain the performance potential of the learned algorithm.

Another approach for improving safety involves using expert-crafted cost functions to optimize the planned output trajectory at test time~\cite{hu2023uniad, Hu2023hoplan}. This method relies on explicit representations, such as drivable area maps or vehicle motion predictions, and defines numerical costs for unsafe planning behaviors in relation to these explicit representations, e.g. driving off-road or collisions. Unfortunately, the representations needed for this approach are costly to annotate and prevent fully leveraging the scalability inherent in end-to-end learning, which typically requires easy-to-obtain annotations such as vehicle trajectories or driving actions only~\cite{Prakash2021transfuser, wu2022tcp}. Consequently, end-to-end planners are often not amenable to test-time optimization using expert-designed cost functions. To safely deploy and improve end-to-end driving policies, we seek a data-driven approach that enhances safety without depending on fallback layers or explicit representations, and that can be effectively scaled.

In this work, we propose such an approach, Centaur, which enhances safety by Test-Time Training (TTT)~\cite{sun2024learning,Liang2023ACS}. This involves computing a gradient of the network weights to minimize an objective function defined using a small dataset collected during deployment~\cite{wang2022continual, liang2024comprehensive, Gao2024fstta}. In contrast to test-time trajectory optimization utilizing explicit representations and expert-designed cost functions, our TTT objective optimizes the network's weights to reduce planning uncertainty, which can be estimated in an unsupervised manner. Furthermore, although we perform test-time training, Centaur performs gradient computations and inference asynchronously and in parallel, leading to a minimal overhead in terms of runtime latency.

Given the historically prevalent paradigm of trajectory regression in end-to-end driving~\cite{hu2023uniad, chitta2023transfuser, Weng2024para, li2024ego}, where the planner directly regresses desired trajectories from sensor inputs, defining a planning uncertainty metric to enable TTT is non-trivial. This is because existing work on TTT focuses on classification~\cite{Gao2024fstta,wang2020tent}, and typically, the training objective used is the prediction's entropy, which is not easy to compute in a regression setting. However, an emerging class of methods in the recent end-to-end driving literature based on trajectory scoring~\cite{li2024hydra, Sadat2020P3, Casas2021mp3} assign numerical scores to multiple trajectory candidates, akin to classifiers. Focusing on this class of methods, we introduce a novel uncertainty measure for end-to-end driving, \uncname. It aggregates candidate trajectory scores from the planner into clusters based on their driving direction, and computes the entropy over this simple and interpretable distribution.

Our approach is exemplified using a trajectory scoring planner, Hydra-MDP~\cite{li2024hydra}, in \figref{fig:main-fig}. In the complex scene depicted involving entering a roundabout, the distribution of high-scoring trajectories becomes more concentrated towards the safer behavior of a right turn after TTT, helping the planner avoid illegally turning left into oncoming traffic. Besides its effectiveness for TTT, our experiments also demonstrate that \uncname can identify situations where the model will fail, suggesting its potential as a mechanism for alerting human drivers to intervene.

Centaur sets a new state-of-the-art ($92.6\%$) on the \texttt{navtest} benchmark, significantly outperforming the conservative fallback layer strategy ($65.3\%$) and closely approaching the upper-bound human performance ($94.8\%$). To advance the progress in safe autonomous driving, we construct \texttt{navsafe}, a new benchmark containing various edge cases and challenging driving scenarios. To achieve this, we devise a semi-automated data curation pipeline involving human annotation and verification to sample safety-critical scenarios. We evaluate the generalization capabilities of state-of-the-art planners, including Centaur, on \texttt{navsafe}, revealing several new insights on their strengths and weaknesses, as well as a substantial gap towards achieving human-level performance in this challenging setting.

\begin{figure*}[t!]
    \centering
    \includegraphics[width=\linewidth]{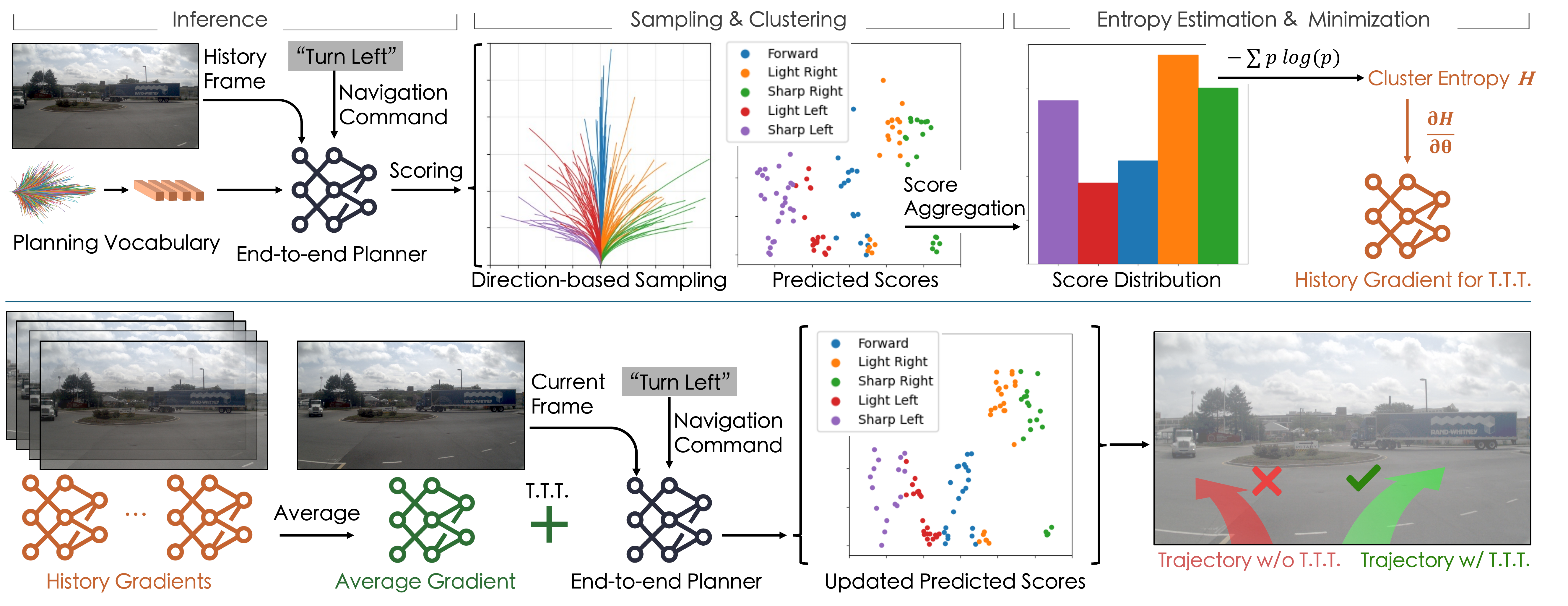}
    \caption{\textbf{Test-Time Training (TTT) in Centaur.} \textbf{Top:} A trained end-to-end planner scores trajectories from the planning vocabulary for frames observed {during testing}. We sample a subset of these, clustered based on their driving direction. After aggregating predicted scores over clusters, a \uncname is calculated to reflect the uncertainty. We then obtain a gradient for \uncname minimization via backpropagation. \textbf{Bottom:} We accumulate gradients from historical frames and update our planner to achieve improved performance.}
    \label{fig:pipeline}
\end{figure*}

\boldparagraph{Contributions} (1) Our work is the first to demonstrate the potential of test-time training for significantly improving performance in end-to-end autonomous driving. (2) We propose \uncname, a simple, interpretable, and effective formulation for measuring uncertainty in end-to-end planning, as evidenced by comprehensive experiments. (3) We release the \texttt{navsafe} benchmark, a collection of safety-critical scenarios such as roundabouts, adverse weather conditions, and unprotected turns. As our approach draws closer to human-level performance on existing benchmarks, this provides a challenging new testbed for future work. 

\section{Centaur}

In this section, 
we present the pipeline of \textbf{\algname}: \textbf{\underline{C}}luster \underline{\textbf{en}}tropy for \textbf{\underline{t}}est-time tr\textbf{\underline{a}}ining using \textbf{\underline{u}}nce\textbf{\underline{r}}tainty. Our work focuses on deploying a trained planner, as shown in~\figref{fig:pipeline}. 

\boldparagraph{Preliminaries} End-to-end self-driving methods seek to obtain a policy, denoted as $\pi$, that directly outputs driving decisions, such as desired trajectories $\bt$, based on sensor inputs $\bx$ and a navigation command $\bc$. Trajectories are of the form $\Tilde{\bt} = (\bp_0, \bp_1, \cdots , \bp_N)$, which is a sequence of waypoints. These are Bird's Eye View (BEV) positions of the vehicle over $N$ evenly spaced time steps. The policy estimates trajectories from sensor inputs $\bx_0$ (such as camera images), and a discrete navigation command $\bc_0$ (indicating driver intention in ambiguous situations, e.g. left/ right/ straight), captured at the current time step, indexed with $0$. Among the existing paradigms for end-to-end driving, the straightforward approach of regressing trajectories using a dataset of human driving demonstrations can be referred to as trajectory regression. While our approach is compatible with trajectory regression, in the following, we primarily address its application to a new emerging paradigm with higher performance in the recent literature: trajectory scoring (\eg, Hydra-MDP~\cite{li2024hydra}, VADv2~\cite{chen2024vadv2} and RAD~\cite{gao2025rad}). We discuss the methodology and results for applying our approach to trajectory regression methods such as UniAD~\cite{hu2023uniad} and TransFuser~\cite{chitta2023transfuser} in the supplementary material.

\boldparagraph{Trajectory Scoring} In this formulation, the planner outputs a set of scalar `score features' $\{s_j\}_{j=1}^k$ for a `planning vocabulary' of $k$ different trajectories. The score features numerically capture different aspects of desired driving behavior, such as avoiding collisions and making progress. These features can then be aggregated into a single final score via an aggregation function $\phi$. During inference, the trajectory with the largest predicted final score is selected:
\begin{equation}
    \label{eqn:multi-score-output}
    \{s_j\}_{j=1}^k  = \{ \pi_\theta(\bt_j, \bx_0, \bc_0) \}_{j=1}^k,
\end{equation}
\begin{equation}
    \label{eqn:cost-select}
    \hat{\bt}_0=\underset{\bt_j}{\arg \max } \{\phi(s_j)\}_{j=1}^k.
\end{equation}
As a result of this design, the planner is inherently multi-modal, with the distribution of scores over the planning vocabulary providing a way to represent planning uncertainty. For this type of planner, learning the policy's parameters $\theta$ can be formulated as:
\begin{equation}
    \label{eqn:objective-score}
   \arg \min_{\theta} \mathbb{E}_{(\bt,\bx,\bc)\sim D_{kd}} [\mathcal{L}_{kd}(e(\bt,\bx,\bc), \pi_\theta(\bt,\bx,\bc))],
\end{equation}
where $e$ is an expert algorithm that can provide ground truth score features for any sampled trajectory, and $D_{kd} = \{(\bt, \bx, \bc), \cdots \}$ is a knowledge distillation dataset generated by sampling this expert algorithm for different inputs, typically in a driving simulator~\cite{caesar2021nuplan, Dosovitskiy17carla} or world model~\cite{yang2024generalized, gao2024vista}. In practice, the score features are normalized to the range $(0,1)$, and $\mathcal{L}_{kd}$ is implemented as a cross-entropy loss.
\subsection{\uncname}
In essence, the scoring procedure from \eqnref{eqn:multi-score-output} and \eqnref{eqn:cost-select} acts as a classifier over the $k$ classes in the planning vocabulary. We formulate a new uncertainty measure, \uncname, to translate this intuition into an unsupervised training objective for end-to-end planning, which does not require an expert algorithm as in \eqnref{eqn:objective-score}. Our key challenge is that the planning vocabulary $k$ is often very large, typically encompassing several thousands of possible trajectories. Therefore, even for simple driving scenarios, several neighboring trajectories in the vocabulary will result in nearly identical scores. Simply minimizing the entropy in the distribution over scores over would result in an objective that encourages the model to avoid trajectories that have a larger number of nearby neighbors in the vocabulary, \eg, driving straight. Therefore, such a naive entropy is unsuitable for uncertainty estimation in this setting.

\boldparagraph{Trajectory Candidates} To tackle this, we first sample a set of $M$ trajectory candidates $\{\bt_m\}_{m=1}^M$ from the planning vocabulary, where $M<<k$. We use a weighted sampling process which prioritizes candidates that have a high average score on the test set as per the expert algorithm $e$. Additional details regarding this procedure are provided in the supplementary material. While the entropy over this reduced set of $M$ scores already serves as a meaningful uncertainty measure, as evidenced by our experiments, we now aim to further improve its effectiveness and interpretability.

\begin{figure*}[ht!]
    \centering
    \includegraphics[width=\linewidth]{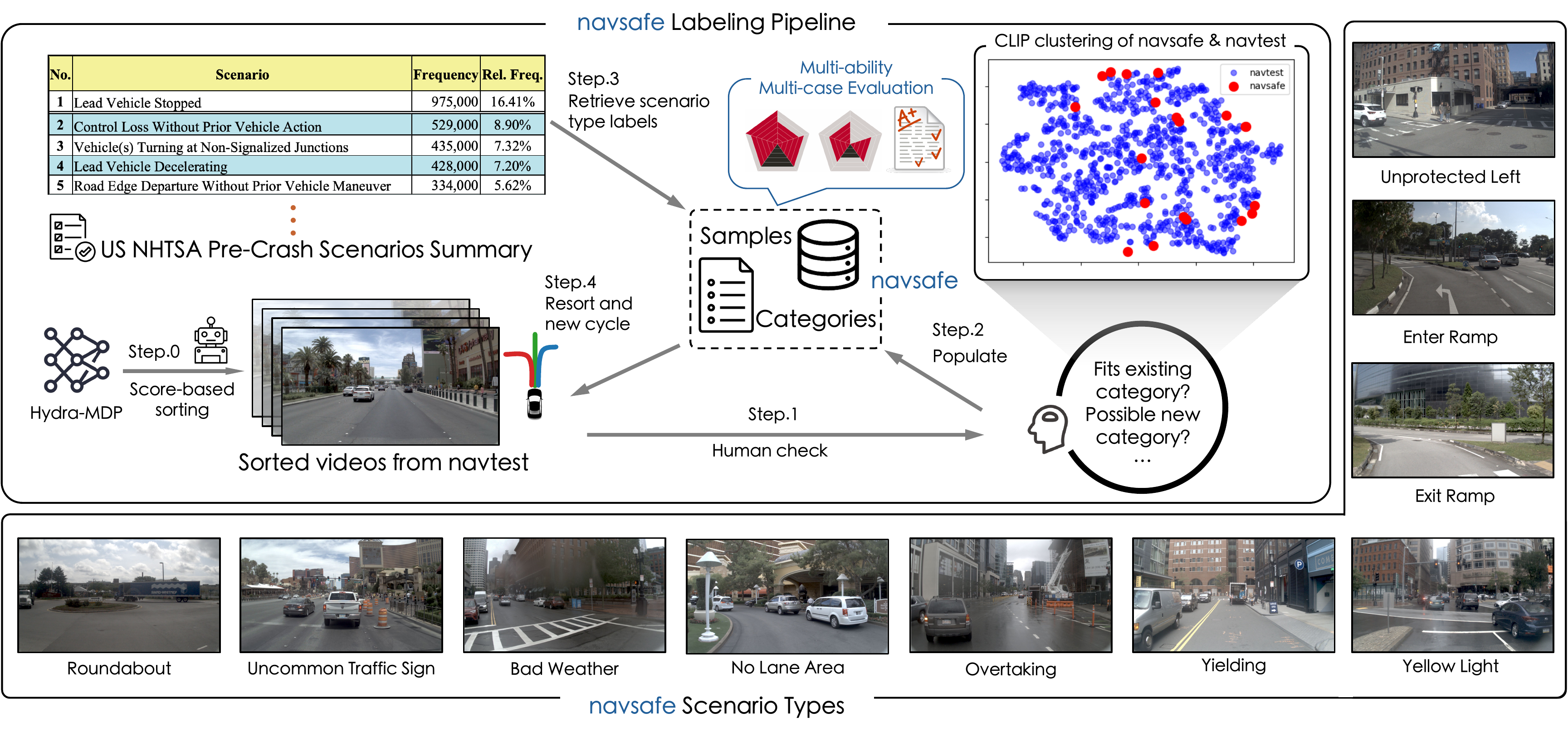}
    \caption{\textbf{\texttt{navsafe}.} We leverage definitions of safety-critical cases from NHTSA (National Highway Traffic Safety Administration) material~\cite{najm2007pre} and search through \texttt{navtest} for frames matching these. Following human checks, we construct \texttt{navsafe}. CLIP-based clustering~\cite{ilharco2021openclip} of our data alongside \texttt{navtest} shows that \texttt{navsafe} consists of frames from the peripheral regions of the distribution.}
    \label{fig:navsafe}
\end{figure*}

\boldparagraph{Direction-based Sampling} Next, we use domain-specific criteria to select trajectory anchors from the $M$ trajectory candidates. Following~\cite{sima2023drivelm}, we select 5 anchors (\texttt{sharp left}, \texttt{slight left}, \texttt{forward}, \texttt{slight right}, \texttt{sharp right}). \texttt{sharp left/right}, the candidates with the largest lateral offsets, are selected first. \texttt{slight left/right} are the candidates with lateral offsets closest to half that of \texttt{sharp left/right}, and \texttt{forward} is the trajectory with the smallest lateral offset. 

\boldparagraph{Score Aggregation} As shown in \figref{fig:pipeline} (top), we aim to construct a 5-way categorical distribution capturing the distribution of predicted scores over the 5 driving directions. For this, we construct 5 clusters by identifying the nearest anchor to each of the $M$ candidates in terms of $L_2$ distance. In each resulting cluster, we add up the predicted scores for all trajectories to obtain an anchor score $score_a$. We compute a probability distribution via normalization ($\frac{score_a}{\sum_{a=1}^5 score_a}$), and define \uncname (denoted as $H$) to be the Shannon Entropy of this distribution.
\subsection{Test-Time Training (TTT)}
Equipped with our unsupervised training objective, our goal is to minimize test-time uncertainty, which is the key idea of TTT approaches in the literature. We follow the standard practice of performing a single step of gradient descent to update our network's parameters $\theta$~\cite{Gao2024fstta}. We use the \uncname $H(\bx)$ as an objective function to optimize. Specifically, given $H(\bx)$ calculated following a forward pass through $\pi_\theta$, we obtain the gradient $\frac{\partial H}{\partial \theta}$ by backpropagation. However, instead of directly updating $\theta$ with the calculated gradient, we store $\frac{\partial H}{\partial \theta}$ in a buffer of length $F=4$, as shown in \figref{fig:pipeline} (bottom). At time-step $i$, we update the network using the average of all gradients $\{\frac{\partial H}{\partial \theta}\}_{avg}$ currently in our buffer, obtaining $\hat{\theta}_i=\theta-\eta \{\frac{\partial H}{\partial \theta}\}_{avg}$, where $\eta$ is the learning rate. Intuitively, this suppresses the model's ability to assign high scores in multiple clusters. This, in turn, corrects the scores of outlier predictions that vary significantly from other trajectories in their cluster.

\boldparagraph{Deployment} Note that the gradient calculated at time-step $i$ only uses gradients of \textit{historical time-steps} already present in the buffer. Therefore, beyond the smoothness provided by gradient averaging, our use of a buffer enables asynchronous and parallel execution. During deployment, the gradient calculation process can be executed on independent computational hardware, so that the more critical inference process 
holds
a minimal overhead in latency. 

\section{\texttt{navsafe}: A Safety-Critical Scenario Set}
\label{sec:navsafe}
In the following, we discuss the construction and highlights of \texttt{navsafe}, a challenging new benchmark consisting of edge cases for end-to-end planning. The data used in our benchmark is sampled from NAVSIM~\cite{dauner2024navsim}, which in turn builds upon data from OpenScene~\cite{openscene2023} and nuPlan~\cite{caesar2021nuplan}. 

\boldparagraph{NAVSIM} The framework on which we build \texttt{navsafe} is primarily designed to capture scenarios of significant driving behavior changes, where the historical trajectory of the ego vehicle does not allow for accurate extrapolation of the future plan. It is split into two parts: \texttt{navtrain} and \texttt{navtest}, which contain 1,192 logs (103,288 frames) and 136 logs (12,146 frames) respectively. In NAVSIM, driving agents are required to plan a trajectory as a sequence of future positions, spanning a horizon of 4 seconds. The input for an agent includes 1.5 seconds of past history (resulting in 4 frames of data) from onboard camera and LiDAR sensors as well as data on the vehicle’s current speed, acceleration, and navigation command, collectively referred to as the ego status. We retain these default NAVSIM settings for evaluations on the \texttt{navsafe} benchmark.

\boldparagraph{Construction Criteria} \figref{fig:navsafe} illustrates the dataset curation process for \texttt{navsafe}. The US NHTSA agency holds a collection of reported traffic accidents every year across the nation, which is shared publicly via a pre-crash scenario typology for crash avoidance research~\cite{najm2007pre}. This includes a wide variety of pre-crash scenario types, ranging from common types such as ``Lead Vehicle Stopped'' to uncommon types such as ``Animal Crash With Prior Vehicle Maneuver''. Supported by the summary statistics on the types of pre-crash scenarios~\cite{thorn2018framework,najm2007pre}, \texttt{navsafe} is curated to highlight these safety-critical scenarios where possible failure cases of modern end-to-end planners could happen. Thus, we follow these steps to construct \texttt{navsafe}:
\begin{enumerate}\addtocounter{enumi}{-1}
    \item Sort the video clips in the \texttt{navtest} split based on the average performance of a state-of-the-art planner~\cite{li2024hydra}.
    \item Check the video clips and try to assign frames to new or existing categories in \texttt{navsafe} with human annotators.
    \item Populate the scenario types with new examples 
    \item Retrieve scenario type labels from the NHTSA file to define scenario types not yet in \texttt{navsafe}.
    \item Resort \texttt{navtest} and start a new cycle of the workflow.
\end{enumerate}

\boldparagraph{Highlights of \texttt{navsafe}} For our benchmark, we select 229 out of 12,146 frames from \texttt{navtest}, covering 10 categories: ``Roundabout'', ``Yellow light, rush or wait?'', ``Exit ramp'', ``Unprotected left'', ``Enter ramp'', ``Uncommon traffic sign'', ``Overtaking with lane change'', ``No lane area (e.g. parking lot)'', ``Bad weather'' and ``Yielding (e.g. to pedestrians)''. By providing independent scores for each scenario type, we enable a fine-grained assesment of specific planning capabilities, unlike the overall aggregate performance which is provided by existing benchmarks like \texttt{navtest}. Our CLIP-based clustering result shows that \texttt{navsafe} represents data points far from the center of the distribution of the original \texttt{navtest} split. In our experiments, we benchmark the human agent performance (ground truth trajectories from NAVSIM) as well as state-of-the-art end-to-end planners to demonstrate the increased level of difficulty posed by this new benchmark. 

\begin{table*}[t!]
    \centering
    \small
    \caption{\textbf{Impact of TTT on \texttt{navtest}.} Introducing a fallback layer~\cite{vitelli2022safetynet} leads to extremely conservative behaviors, in turn reducing PDMS. On the other hand, TTT yields significant improvements, in particular with our proposed Cluster Entropy uncertainty measure.}
    \label{tab:navtest_ablation}
    \setlength{\tabcolsep}{2.7mm}
    \begin{tabular}{l|ll|rrrr|rr}
    \toprule
    \textbf{Model} & \textbf{Deployment} & \textbf{Uncertainty} & \textbf{NC} $\uparrow$ & \textbf{DAC} $\uparrow$ & \textbf{EP} $\uparrow$ & \textbf{C} $\uparrow$ & \textbf{TTC} $\uparrow$ & \textbf{PDMS} $\uparrow$  \\
    \midrule
    \multirow{7}{*}{Hydra-MDP~\cite{li2024hydra}} & - & - & $98.4$ & $97.8$ & $\mathbf{86.5}$ & ${100.0}$ & $93.9$ & $90.3$  \\
    \cmidrule(lr){2-9}
    & \multirow{4}{*}{Fallback Layer~\cite{vitelli2022safetynet}} & KL Divergence & $99.1$ & $99.2$ & $9.6$ & ${100.0}$ & $96.7$ & $63.1$  \\
    & & Full Entropy & $99.6$ & $ 99.4$ & $13.2$ & ${100.0}$ & $99.1$ & $65.9$  \\
    & & Semantic Entropy & $\mathbf{99.7}$ & $\mathbf{99.5}$ & $10.4$ & ${100.0}$ & $98.9$ & $65.3$  \\
    & & Cluster Entropy & $\mathbf{99.7}$ & $\mathbf{99.5}$ & $8.2$ & ${100.0}$ & $\mathbf{99.3}$ & $64.2$   \\
    \cmidrule(lr){2-9}
    & \multirow{2}{*}{TTT} & KL Divergence & $98.9$ & $98.1$ & $86.0$ & ${100.0}$ & $94.4$ & $91.5$  \\
    & & Full Entropy & $99.3$ & $ 98.5$ & $85.7$ & ${100.0}$ & $97.1$ & $91.8$   \\
    \midrule
    Hydra-SE & TTT & Semantic Entropy & $99.2$ & $98.5$ & $85.7$ & ${100.0}$ & $97.1$ & $91.8$   \\ 
    \textbf{Centaur (Ours)} & TTT & Cluster Entropy & $99.5$ & $98.9$ & $85.9$ & ${100.0}$ & $98.0$ & $\mathbf{92.6}$   \\
    \midrule
    \textit{Human Trajectory} & - & - & $\mathit{100.0}$ & $\mathit{100.0}$ & $\mathit{87.5}$ & $\mathit{99.9}$ & $\mathit{100.0}$ & $\mathit{94.8}$ \\
    \bottomrule
    \end{tabular}
\end{table*}

\begin{table*}[t!]
    \centering
    \caption{\textbf{\texttt{navtest} Leaderboard.} We report the mean and std over 3 independent training runs, as recommended in the leaderboard rules. 
    *The Hydra-MDP entry, which won the 2024 NAVSIM challenge, is a single evaluation of a 3-model ensemble, unlike all other rows. 
    }
    \label{tab:navtest_all}
    \small %
    \begin{tabular}{l|llll|ll}
    \toprule
    \textbf{Model} & \textbf{NC} $\uparrow$ & \textbf{DAC} $\uparrow$ & \textbf{EP} $\uparrow$ & \textbf{C} $\uparrow$ & \textbf{TTC} $\uparrow$ & \textbf{PDMS} $\uparrow$ \\
    \midrule
    \textbf{Centaur (Ours)} & $99.23\pm0.35$ & $\mathbf{98.72\pm0.22}$ & $\mathbf{85.96\pm0.77}$ & $99.97\pm0.03$ & $97.17\pm1.06$ & $\mathbf{92.10\pm0.33}$ \\ 
    Hydra-SE & $\mathbf{99.32\pm0.05}$ & ${98.58\pm0.12}$ & ${85.42\pm0.07}$ & $99.96\pm0.01$ & $\mathbf{97.30\pm0.23}$ & ${91.87\pm0.19}$ \\
    \midrule
    Hydra-MDP*~\cite{li2024hydra} & $99.07$ & $98.29$ & $85.20$ & $\mathbf{100.0}$ & $96.56$ & $91.26$  \\
    TransFuser~\citep{chitta2023transfuser} & $97.78\pm0.10$ & $92.63\pm0.48$ & $78.88\pm0.25$ & $99.98\pm0.01$ & $92.89\pm0.21$ & $83.88 \pm 0.45$\\
    LTF~\citep{chitta2023transfuser} & $97.68 \pm 0.11$ & $92.29 \pm 0.45$ & $78.33 \pm 0.53$ & $99.99 \pm 0.01$ & $93.10 \pm 0.19$ & $83.52 \pm 0.55$\\
    Ego Status MLP~\cite{dauner2024navsim} & $93.09 \pm 0.12$ & $78.26 \pm 0.98$ & $63.20 \pm 0.53$ & $99.97 \pm 0.02$ & $84.02 \pm 0.61$ & $66.40 \pm 0.94$\\
    \bottomrule
    \end{tabular}
\end{table*}

\section{Experiments}
In this section, we present the performance of Centaur on the NAVSIM framework~\cite{dauner2024navsim}, including the official leaderboard (\texttt{navtest}) and our proposed \texttt{navsafe} safety-critical scenarios, as detailed in \secref{sec:navsafe}. 

\boldparagraph{Metrics} Our experiments adopt the primary metric of NAVSIM, PDM score (PDMS), to evaluate the performance of end-to-end planning models:
\begin{equation}\label{eq:pdms}
\mathrm{PDMS} = \mathrm{NC} \times \mathrm{DAC} \times \left( \frac{5 \mathrm{TTC} + 2 \mathrm{C} + 5 \mathrm{EP} }{12} \right),
\end{equation}
where the sub-scores \textbf{NC} (No
at-fault Collision), \textbf{DAC} (Drivable Area Compliance), \textbf{EP} (Ego Progress), \textbf{C} (Comfort), and \textbf{TTC} (Time-to-Collision),  each represented as a percentage, are composed into a single score. A detailed definition of each subscore can be found in~\cite{dauner2024navsim}. Due to its relevance for safety, in several experiments, we discuss TTC in particular, which identifies whether the planner's output trajectory results in near-collisions, i.e., a forward projection at a constant velocity would result in a collision within 1 second at any point along the trajectory.

\subsection{Baselines}
For our experiments, we consider several baseline approaches to analyze the impact of various design choices.

\boldparagraph{Base Model} Designed specifically for NAVSIM, our base model Hydra-MDP~\cite{li2024hydra} 
inputs LiDAR points which are splatted onto the BEV plane and encoded via a ResNet34~\cite{he2016deep}. These features are fused with image-based features extracted using a V2-99~\cite{park2021pseudo} backbone via TransFuser~\cite{chitta2023transfuser} sensor fusion modules. The decoder then predicts scores for a planning vocabulary of $k=8192$ trajectories. The score functions output by the decoder are the five sub-scores (NC, DAC, EP, C, TTC) considered during NAVSIM evaluation. 
Additional experiments using the more lightweight TransFuser~\cite{chitta2023transfuser} baseline,  which performs simple trajectory regression instead of scoring like Hydra-MDP, are provided in the supplementary material.

\boldparagraph{Deployment Strategy} As a baseline for safer deployment, we implement a fallback layer strategy following the description of~\cite{vitelli2022safetynet}. We first calculate the average PDMS of each trajectory in the planning vocabulary of Hydra-MDP and select top 20 of them as fallback trajectories. During inference, we measure the $L_2$ distance between the predicted trajectory and the 20 fallback trajectories. The closest among these 20 fallback trajectories is selected as the safe action to replace the predicted one, in case a failure is detected based on an uncertainty value above a threshold. The thresholds we choose are later discussed in \tabref{tab:fail_class}.

\boldparagraph{Uncertainty Measure} We compare Cluster Entropy to three baselines. For \textbf{KL Divergence}, we obtain the predicted categorical distribution over planning vocabulary ($k=8192$ classes) per score function (NC, DAC, EP, C, TTC). We then calculate KL-divergence between each pair of distributions and sum over all pairs.
For \textbf{Full Entropy}~\cite{shannon1948entropy}, our calculation is similar to \uncname. However, we skip the step of clustering the $M=100$ candidates into 5 clusters, and simply compute the entropy over the full 100-way candidate score distribution.
\textbf{Semantic Entropy} is our adaptation of~\cite{kuhn2023semantic} for autonomous driving. Again, the calculation is similar to \uncname, but we cluster candidates using a feature space where each trajectory is represented by its predicted scores $s_j$. More details are provided in the supplementary material. %
For compactness, we refer to the approach of TTT for Hydra-MDP with Semantic Entropy, which is the most frequently used baseline in our experiments, as \textbf{Hydra-SE}. 

\begin{figure*}[t]
    \centering
    \includegraphics[width=\linewidth]{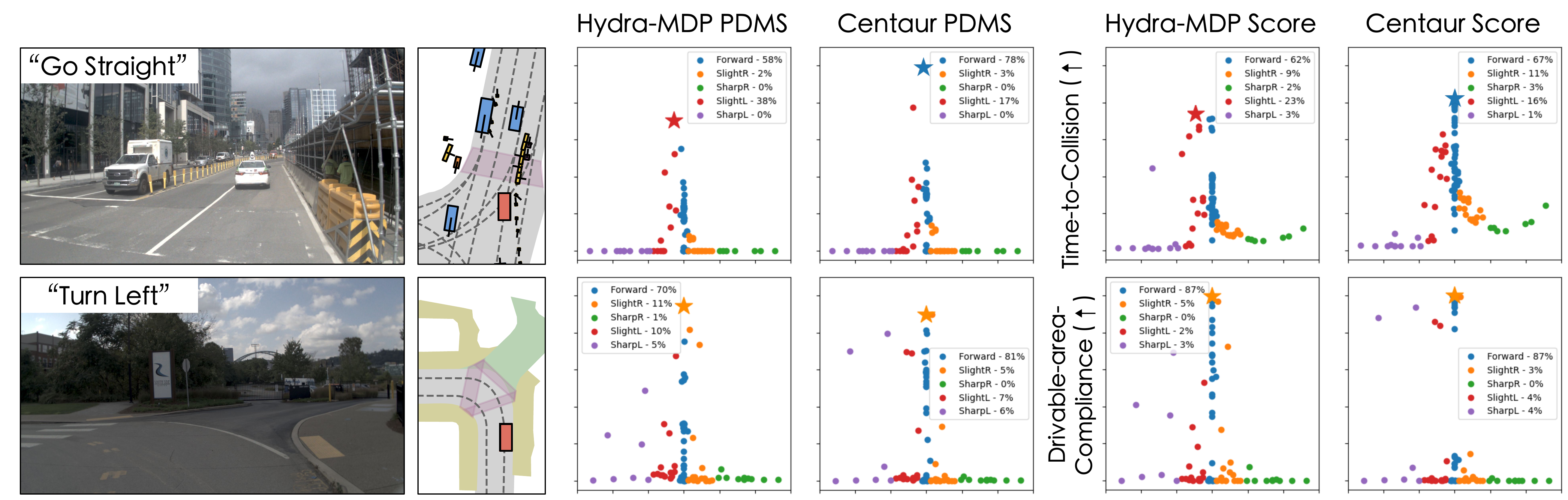}
    \caption{\textbf{Qualitative results.} For each scene, we show the PDMS and a selected subscore for both Hydra-MDP (before TTT) and Centaur (after TTT), with the highest predicted score marked using a $\filledstar$. The x-axis in each plot is the candidate trajectory's lateral end-point position. \textbf{Top:} TTT helps Centaur which is uncertain between `Forward' and `Slight Left' to prefer the direction where cluster members have a higher average score, improving safety. \textbf{Bottom:} A failure case, where TTT cannot suppress a confident original prediction.}
    \label{fig:vis}
\end{figure*}

\boldparagraph{Implementation} Except for \tabref{tab:navtest_all}, we adopt NAVSIM v1.0 for all the training and local evaluations, and NAVSIM v1.1 for our official leaderboard submissions, between which slight differences arise from minor changes in metric definitions. We train the base end-to-end model Hydra-MDP~\cite{li2024hydra} on the \texttt{navtrain} split for 20 epochs using 8 NVIDIA A100 GPUs, with a total batch size of 256. The AdamW~\cite{loshchilov2017decoupled} learning rate is set to $10^{-4}$ with weight decay disabled. For \uncname calculation and TTT, we use $M=100$ candidate trajectories and perform 1 step of gradient descent with a learning rate $\eta$ of $10^{-4}$ on a dataset of $F=4$ history frames. We only update the score decoder during TTT, keeping the perception backbones frozen.

\subsection{Results}

Our main results are shown in \tabref{tab:navtest_ablation}, examining two key aspects: deployment strategy and uncertainty measurement.

\boldparagraph{Deployment Strategy} Fallback layers, irrespective of the uncertainty measure employed, notably decrease the PDM Score. This decline can be primarily attributed to a considerable reduction in ego progress, from 86.5 to below 15. Unfortunately, the more failure cases the model identifies, the worse the fallback layer PDMS is due to the conservative fallback trajectories. Although the safety-related sub-scores (NC, DAC and TTC) improve, this approach would likely lead to an over-reliance on human intervention and be infeasible in practice. In contrast, TTT improves the safety-related sub-scores while maintaining high progress. The most significant increase is observed in TTC, highlighting its potential for improved safety. %

\begin{table}[t!]
    \centering
    \caption{\textbf{Failure identification.} Our proposed Cluster Entropy uncertainty with Hydra-MDP obtains promising results on identifying frames where the model will fail on \texttt{navtest}.}
    \label{tab:fail_class}
    \small %
    \begin{tabular}{l|c|rr}
    \toprule
    \textbf{Uncertainty} & \textbf{Threshold} & \textbf{TPR} $\uparrow$ & \textbf{Acc.} $\uparrow$
    \\
    \midrule
    KL Divergence  & $1500$ & $41.3$ & $68.3$ \\
    Full Entropy & $0.8$ & $61.7$ & $70.4$ \\
    Semantic Entropy & $0.8$ & ${62.5}$ & ${72.2}$ \\
    \textbf{Cluster Entropy (Ours)} & $0.8$ & $\mathbf{62.8}$ & $\mathbf{73.6}$ \\
    \midrule
    \textit{Select All} & - & $\mathit{100.0}$ & $\mathit{15.6}$ \\
    \textit{Select None}  & - & $\mathit{0.0}$ & $\mathit{84.4}$ \\
    \bottomrule
    \end{tabular}
    \vspace{-10pt}
\end{table}
\begin{table*}[t!]
    \centering
    \caption{\textbf{\texttt{navsafe} PDM Scores.} Existing planners fall short of human-level performance. \textbf{RDBT}: ``Roundabout''. \textbf{YLLT}: ``Yellow light, rush or wait?''. \textbf{EXR}: ``Exit ramp''. \textbf{UNPL}: ``Unprotected left''. \textbf{ENR}: ``Enter ramp''. \textbf{UNTS}: ``Uncommon traffic sign''. \textbf{OTLC}: ``Overtaking with lane change''. \textbf{NLA}: ``No lane area (e.g., parking lot)''. \textbf{BWTH}: ``Bad weather''. \textbf{YLD}: ``Yielding (e.g., to pedestrians)''.}
    \label{tab:navsafe}
    \small %
    \setlength{\tabcolsep}{2.2mm}
    \begin{tabular}{l|rrrrrrrrrr|r}
    \toprule
    & \textbf{RDBT} & \textbf{YLLT} & \textbf{EXR} & \textbf{UNPL} & \textbf{ENR} & \textbf{UNTS} & \textbf{OTLC} & \textbf{NLA} & \textbf{BWTH} & \textbf{YLD} & \textbf{Overall} \\
    \midrule
    Num. Scenarios & $7$ & $23$ & $45$ & $28$ & $51$ & $12$ & $8$ & $5$ & $39$ & $11$ & $229$ \\
    \midrule
    Constant Velocity~\cite{dauner2024navsim} & $0.0$ & $3.8$ & $3.6$ & $0.0$ & $2.9$ & $7.4$ & $0.0$ & $6.4$ & $6.3$ & $3.8$ & $3.41$ \\
    Ego Status MLP~\cite{dauner2024navsim} & $28.9$ & $22.7$ & $26.0$ & $8.9$ & $37.7$ & $28.0$ & $35.0$ & $8.8$ & $33.4$ & $22.1$ & $24.10$ \\
    LTF~\cite{chitta2023transfuser} & $42.8$ & $\mathbf{70.2}$ & $43.8$ & $59.4$ & $70.6$ & $55.3$ & $55.8$ & $31.5$ & $62.8$ & $47.5$ & $53.74$ \\
    TransFuser~\cite{chitta2023transfuser} & $47.3$ & $55.1$ & $35.6$ & $55.7$ & $66.3$ & $51.7$ & $56.2$ & $42.7$ & $55.2$ & $61.6$ & $51.87$ \\
    Hydra-MDP~\cite{li2024hydra} & $32.6$ & $34.3$ & $37.7$ & ${64.9}$ & $65.5$ & $59.6$ & $43.3$ & $38.1$ & $35.7$ & $71.6$ & $56.47$ \\
    \midrule
    Hydra-SE & $62.2$ & $53.9$ & $50.5$ & $57.8$ & $72.0$ & $65.6$ & $74.0$ & $57.6$ & $\mathbf{64.4}$ & $\mathbf{84.6}$ & $62.84$ \\
    \textbf{Centaur (Ours)} & $\mathbf{76.5}$ & $65.0$ & $\mathbf{61.4}$ & $\mathbf{73.6}$ & $\mathbf{75.0}$ & $\mathbf{85.2}$ & $\mathbf{83.6}$ & $\mathbf{79.8}$ & $60.2$ & $59.3$ & $\mathbf{74.14}$ \\
    \midrule
    \textit{Human Trajectory} & $\mathit{96.4}$ & $\mathit{97.6}$ & $\mathit{95.2}$ & $\mathit{94.3}$ & $\mathit{97.2}$ & $\mathit{95.8}$ & $\mathit{94.1}$ & $\mathit{93.7}$ & $\mathit{90.7}$ & $\mathit{91.5}$ & $\mathit{94.65}$ \\
    \bottomrule
    \end{tabular}
\end{table*}

\boldparagraph{Uncertainty Measure} When using TTT with Cluster Entropy, our improvements of 2.3 PDMS for Centaur compared to the prior state-of-the-art Hydra-MDP and 0.6 PDMS compared to the Semantic Entropy based method Hydra-SE baseline are substantial. This is because scores on \texttt{navtest} are already very close to the upper bound human PDMS of 94.8. Besides the quantitative gains, to demonstrate how Cluster Entropy facilitates interpretability through its intuitive clustering mechanism, we visualize two scenarios in \figref{fig:vis}. In the top row, the base Hydra-MDP model without TTT shows a preference for a slight left trajectory, which does not maintain a large time-to-collision margin. After TTT, Centaur follows the safer behavior of continuing forward. In the bottom row, TTT raises the Drivable Area Compliance score of two `Sharp Left' candidates, but this is insufficient to raise their score above the original prediction. As a result, there is no change in the model's prediction for this frame after TTT.

\boldparagraph{Failure Identification} In this experiment, we perform binary classification of potential failure cases using different uncertainty estimates. This task is of critical relevance for vehicles deployed with partial autonomy which rely on human intervention in challenging situations. We define a failure as a scenario where the model obtains a PDMS of 0, and report the true positive rates (TPR) and accuracies for identifying such failures in \tabref{tab:fail_class}. We consider uncertainty above a threshold to indicate a possible failure. From our results, we see that Cluster Entropy shows the most promise, outperforming all other uncertainty-based failure classifiers, with a TPR of 62.8\%. Here, we select a threshold of $0.8=0.5\ln5$, which is half the entropy value of a uniform distribution over 5 classes. Based on these findings, we focus on the two best uncertainty measures (Semantic Entropy and Cluster Entropy) in the following experiments. Interestingly, when using TTT, we observe that the PDMS a reliable indicator of the quality of uncertainty measurement, but not when using a fallback layer. For example, the most effective failure case identifier in \tabref{tab:fail_class} (\ie, Cluster Entropy) also obtains the best PDMS with TTT in \tabref{tab:navtest_ablation}, but triggers the fallback mechanism most often, leading to lower PDMS with a fallback layer than Semantic Entropy. 

\boldparagraph{\texttt{navtest} Leaderboard} Following the official instructions for leaderboard submissions, we train our models with 3 random initializations, and submit Centaur as well as Hydra-SE to the \texttt{navtest} leaderboard~\cite{dauner2024navsim} which uses NAVSIM v1.1. Comparisons to additional baselines using NAVSIM v1.0 are provided in the supplementary material. \tabref{tab:navtest_all} shows the mean and standard deviation of the metrics for all entries on the leaderboard submitted with 3 seeds, sharing the same setting as ours. Centaur ranks first, with a mean PDMS of 92.1\%. We achieve this without any training hyper-parameter tuning beyond the Hydra-MDP defaults, and do not require model ensembling like the prior Hydra-MDP result which is based on the winning entry to the 2024 NAVSIM challenge~\cite{li2024hydra}. 

\boldparagraph{\texttt{navsafe}} Finally, we conduct a scenario-level evaluation of different planners on the proposed \texttt{navsafe} benchmark. As seen in \tabref{tab:navsafe}, our new test set captures common failures across several planners, such as EXR and NLA where all existing approaches (which do not conduct TTT) score lower than 50. Our evaluation protocol also provides insights about unique shortcomings and strengths of certain planners. For example, in the scenarios RDBT and OTLC, Hydra-MDP has a similar score to the naive Ego Status MLP baseline. On the other hand, the camera-only LTF model performs surprisingly well on YLLT and ENR, even when compared to the more powerful sensor fusion models such as TransFuser. Our proposed approach Centaur improves significantly over Hydra-MDP as well as Hydra-SE in these challenging scenarios, showing the effectiveness of Cluster Entropy for TTT. In addition, unlike \texttt{navtest}, there is still a wide gap between Centaur and human-level performance, encouraging further effort on this benchmark.

\section{Related Work}

\boldparagraph{End-to-End Autonomous Driving} End-to-end autonomous driving represents a paradigm shift in AV technology~\cite{chen2023e2esurvey}. The goal of these methods is to directly map raw sensor inputs to vehicle control outputs using neural networks, bypassing traditional modular components such as separate perception, planning, and control systems~\cite{Prakash2021transfuser,chitta2021neat,jaeger2023hidden,zimmerlin2024tf,chen2024vadv2,jiang2023vad,hu2023uniad,hu2022model,jia2023think,wang2023drivedreamer,jia2023driveadapter,li2024hydra,tong2023occnet,liao2025diffusiondrive}.

\boldparagraph{End-to-End Planner Deployment} Obtaining trustworthy outputs from an end-to-end system during deployment is challenging, considering its ``black box'' nature. Rule-based planners or fall-back layers are common ways to guardrail the planning output. For instance, UniAD~\cite{hu2023uniad} uses test-time optimization to refine the planning output and TransFuser~\cite{chitta2023transfuser} adopts forced braking when obstacles are encountered by the LiDAR sensor. SafetyNet~\cite{vitelli2022safetynet} adopts fallback layers to ensure planning safety via sanity checks on collision and physical feasibility.
However, unlike Centaur, these works either require heavy annotation during training time or result in overly conservative actions.

\boldparagraph{Test-time Compute \& Training} Leveraging test-time compute has recently emerged as an important way to strengthen reasoning capabilities~\cite{openaio1}. In addition, test-time adaptation and training techniques~\cite{Liang2023ACS} have shown considerable potential in computer vision~\cite{sun2024learning,hakim2023clust3,marsden2024universal}, multimodal learning~\cite{shu2022test,yoon2024c}, vision-language navigation~\cite{Gao2024fstta}, trajectory prediction~\cite{park2024t4p}, robot manipulation~\cite{hansen2021self} and long context LLMs~\cite{sun2024learning}. The key idea of these approaches is to minimize entropy or related proxies at test time. Centaur is aligned in spirit with these approaches and to the best of our knowledge, it is the first work that explores the potential of test-time training via uncertainty measurement for the vehicle motion planning task.

\section{Conclusion} 
In this work, we make a first attempt towards employing Test-Time Training (TTT) for end-to-end autonomous driving and find that it significantly improves the performance of state-of-the-art methods on the official \texttt{navtest} leaderboard. Our proposed \uncname metric quantifies uncertainty effectively, outperforming several baseline methods. Finally, to provide a holistic evaluation of performance in challenging scenarios, we release the \texttt{navsafe} test split, providing new insights on the key areas of improvement for end-to-end driving algorithms. 

\boldparagraph{Limitations} As a pioneering attempt, Centaur shows promising performance. However, there are certain limitations. Our approach is currently evaluated under an open-loop scheme, albeit with principled, simulation-based metrics. Extending our work to a closed-loop setting with an affordable inference budget is a promising direction to explore. In addition, test-time compute is quite limited for autonomous driving as it is an efficiency-sensitive task. 
It is worth exploring ways of unifying TTT and model inference to 
minimize the compute overhead. A discussion on the latency of Centaur is included in the supplementary material.

\section*{Acknowledgements}
We extend our gratitude to Tianyu Li, Zhenxin Li, Nikita Durasov, Qingwen Bu, Jiwoong Choi and Li Chen for their profound discussions and supportive help in writing. This work is supported by HKU Shanghai Intelligent Computing Research Center, and also partially supported by the EXC (number 2064/1 – project number: 390727645). We thank the International Max Planck Research School for Intelligent Systems (IMPRS-IS) for supporting Kashyap Chitta.

{
    \small
    \bibliographystyle{ieeenat_fullname}
    \bibliography{main}

\begin{thebibliography}{63}
\providecommand{\natexlab}[1]{#1}
\providecommand{\url}[1]{\texttt{#1}}
\expandafter\ifx\csname urlstyle\endcsname\relax
  \providecommand{\doi}[1]{doi: #1}\else
  \providecommand{\doi}{doi: \begingroup \urlstyle{rm}\Url}\fi

\bibitem[Amini et~al.(2020)Amini, Schwarting, Soleimany, and Rus]{amini2020deepevidential}
Alexander Amini, Wilko Schwarting, Ava Soleimany, and Daniela Rus.
\newblock Deep evidential regression.
\newblock In \emph{NeurIPS}, 2020.

\bibitem[Caesar et~al.(2021)Caesar, Kabzan, Tan, Fong, Wolff, Lang, Fletcher, Beijbom, and Omari]{caesar2021nuplan}
Holger Caesar, Juraj Kabzan, Kok~Seang Tan, Whye~Kit Fong, Eric~M. Wolff, Alex~H. Lang, Luke Fletcher, Oscar Beijbom, and Sammy Omari.
\newblock {nuPlan}: {A} closed-loop ml-based planning benchmark for autonomous vehicles.
\newblock In \emph{CVPR Workshops}, 2021.

\bibitem[Casas et~al.(2021)Casas, Sadat, and Urtasun]{Casas2021mp3}
Sergio Casas, Abbas Sadat, and Raquel Urtasun.
\newblock {MP3}: A unified model to map, perceive, predict and plan.
\newblock In \emph{CVPR}, 2021.

\bibitem[Chen et~al.(2024{\natexlab{a}})Chen, Wu, Chitta, Jaeger, Geiger, and Li]{chen2023e2esurvey}
Li Chen, Penghao Wu, Kashyap Chitta, Bernhard Jaeger, Andreas Geiger, and Hongyang Li.
\newblock End-to-end autonomous driving: Challenges and frontiers.
\newblock \emph{IEEE TPAMI}, 2024{\natexlab{a}}.

\bibitem[Chen et~al.(2024{\natexlab{b}})Chen, Jiang, Gao, Liao, Xu, Zhang, Huang, Liu, and Wang]{chen2024vadv2}
Shaoyu Chen, Bo Jiang, Hao Gao, Bencheng Liao, Qing Xu, Qian Zhang, Chang Huang, Wenyu Liu, and Xinggang Wang.
\newblock {VADv2}: End-to-end vectorized autonomous driving via probabilistic planning.
\newblock \emph{arXiv preprint arXiv:2402.13243}, 2024{\natexlab{b}}.

\bibitem[Chitta et~al.(2021)Chitta, Prakash, and Geiger]{chitta2021neat}
Kashyap Chitta, Aditya Prakash, and Andreas Geiger.
\newblock Neat: Neural attention fields for end-to-end autonomous driving.
\newblock In \emph{International Conference on Computer Vision (ICCV)}, 2021.

\bibitem[Chitta et~al.(2023)Chitta, Prakash, Jaeger, Yu, Renz, and Geiger]{chitta2023transfuser}
Kashyap Chitta, Aditya Prakash, Bernhard Jaeger, Zehao Yu, Katrin Renz, and Andreas Geiger.
\newblock {TransFuser}: Imitation with transformer-based sensor fusion for autonomous driving.
\newblock \emph{IEEE TPAMI}, 2023.

\bibitem[Contributors(2023)]{openscene2023}
OpenScene Contributors.
\newblock {OpenScene}: The largest up-to-date 3d occupancy prediction benchmark in autonomous driving.
\newblock \url{https://github.com/OpenDriveLab/OpenScene}, 2023.

\bibitem[Dauner et~al.(2023)Dauner, Hallgarten, Geiger, and Chitta]{dauner2023parting}
Daniel Dauner, Marcel Hallgarten, Andreas Geiger, and Kashyap Chitta.
\newblock Parting with misconceptions about learning-based vehicle motion planning.
\newblock In \emph{CoRL}, 2023.

\bibitem[Dauner et~al.(2024)Dauner, Hallgarten, Li, Weng, Huang, Yang, Li, Gilitschenski, Ivanovic, Pavone, Geiger, and Chitta]{dauner2024navsim}
Daniel Dauner, Marcel Hallgarten, Tianyu Li, Xinshuo Weng, Zhiyu Huang, Zetong Yang, Hongyang Li, Igor Gilitschenski, Boris Ivanovic, Marco Pavone, Andreas Geiger, and Kashyap Chitta.
\newblock {NAVSIM}: Data-driven non-reactive autonomous vehicle simulation and benchmarking.
\newblock In \emph{NeurIPS}, 2024.

\bibitem[Dosovitskiy et~al.(2017)Dosovitskiy, Ros, Codevilla, Lopez, and Koltun]{Dosovitskiy17carla}
Alexey Dosovitskiy, German Ros, Felipe Codevilla, Antonio Lopez, and Vladlen Koltun.
\newblock {CARLA}: {An} open urban driving simulator.
\newblock In \emph{CoRL}, 2017.

\bibitem[Gao et~al.(2025)Gao, Chen, Jiang, Liao, Shi, Guo, Pu, Yin, Li, Zhang, Zhang, Liu, Zhang, and Wang]{gao2025rad}
Hao Gao, Shaoyu Chen, Bo Jiang, Bencheng Liao, Yiang Shi, Xiaoyang Guo, Yuechuan Pu, Haoran Yin, Xiangyu Li, Xinbang Zhang, Ying Zhang, Wenyu Liu, Qian Zhang, and Xinggang Wang.
\newblock Rad: Training an end-to-end driving policy via large-scale 3dgs-based reinforcement learning.
\newblock \emph{arXiv preprint arXiv:2502.13144}, 2025.

\bibitem[Gao et~al.(2024{\natexlab{a}})Gao, Yao, and Xu]{Gao2024fstta}
Junyu Gao, Xuan Yao, and Changsheng Xu.
\newblock Fast-slow test-time adaptation for online vision-and-language navigation.
\newblock In \emph{ICML}, 2024{\natexlab{a}}.

\bibitem[Gao et~al.(2024{\natexlab{b}})Gao, Yang, Chen, Chitta, Qiu, Geiger, Zhang, and Li]{gao2024vista}
Shenyuan Gao, Jiazhi Yang, Li Chen, Kashyap Chitta, Yihang Qiu, Andreas Geiger, Jun Zhang, and Hongyang Li.
\newblock Vista: A generalizable driving world model with high fidelity and versatile controllability.
\newblock 2024{\natexlab{b}}.

\bibitem[Hakim et~al.(2023)Hakim, Osowiechi, Noori, Cheraghalikhani, Bahri, Ben~Ayed, and Desrosiers]{hakim2023clust3}
Gustavo A~Vargas Hakim, David Osowiechi, Mehrdad Noori, Milad Cheraghalikhani, Ali Bahri, Ismail Ben~Ayed, and Christian Desrosiers.
\newblock Clust3: Information invariant test-time training.
\newblock In \emph{ICCV}, 2023.

\bibitem[Hallgarten et~al.(2024)Hallgarten, Zapata, Stoll, Renz, and Zell]{hallgarten2024interplan}
Marcel Hallgarten, Julian Zapata, Martin Stoll, Katrin Renz, and Andreas Zell.
\newblock Can vehicle motion planning generalize to realistic long-tail scenarios?, 2024.

\bibitem[Hansen et~al.(2021)Hansen, Jangir, Sun, Aleny{\`a}, Abbeel, Efros, Pinto, and Wang]{hansen2021self}
Nicklas Hansen, Rishabh Jangir, Yu Sun, Guillem Aleny{\`a}, Pieter Abbeel, Alexei~A Efros, Lerrel Pinto, and Xiaolong Wang.
\newblock Self-supervised policy adaptation during deployment.
\newblock In \emph{ICLR}, 2021.

\bibitem[He et~al.(2016)He, Zhang, Ren, and Sun]{he2016deep}
Kaiming He, Xiangyu Zhang, Shaoqing Ren, and Jian Sun.
\newblock Deep residual learning for image recognition.
\newblock In \emph{CVPR}, 2016.

\bibitem[Hu et~al.(2022)Hu, Corrado, Griffiths, Murez, Gurau, Yeo, Kendall, Cipolla, and Shotton]{hu2022model}
Anthony Hu, Gianluca Corrado, Nicolas Griffiths, Zachary Murez, Corina Gurau, Hudson Yeo, Alex Kendall, Roberto Cipolla, and Jamie Shotton.
\newblock Model-based imitation learning for urban driving.
\newblock In \emph{NeurIPS}, 2022.

\bibitem[Hu et~al.(2023{\natexlab{a}})Hu, Russell, Yeo, Murez, Fedoseev, Kendall, Shotton, and Corrado]{hu2023gaia}
Anthony Hu, Lloyd Russell, Hudson Yeo, Zak Murez, George Fedoseev, Alex Kendall, Jamie Shotton, and Gianluca Corrado.
\newblock {GAIA-1}: A generative world model for autonomous driving.
\newblock \emph{arXiv preprint arXiv:2309.17080}, 2023{\natexlab{a}}.

\bibitem[Hu et~al.(2023{\natexlab{b}})Hu, Li, Liang, Qian, Yang, Zhang, Shao, Ding, Xu, and Liu]{Hu2023hoplan}
Yihan Hu, Kun Li, Pingyuan Liang, Jingyu Qian, Zhening Yang, Haichao Zhang, Wenxin Shao, Zhuangzhuang Ding, Wei Xu, and Qiang Liu.
\newblock Imitation with spatial-temporal heatmap: 2nd place solution for nuplan challenge.
\newblock \emph{arXiv preprint arXiv:2306.15700}, 2023{\natexlab{b}}.

\bibitem[Hu et~al.(2023{\natexlab{c}})Hu, Yang, Chen, Li, Sima, Zhu, Chai, Du, Lin, et~al.]{hu2023uniad}
Yihan Hu, Jiazhi Yang, Li Chen, Keyu Li, Chonghao Sima, Xizhou Zhu, Siqi Chai, Senyao Du, Tianwei Lin, et~al.
\newblock Planning-oriented autonomous driving.
\newblock In \emph{CVPR}, 2023{\natexlab{c}}.

\bibitem[Ilharco et~al.(2021)Ilharco, Wortsman, Wightman, Gordon, Carlini, Taori, Dave, Shankar, Namkoong, Miller, Hajishirzi, Farhadi, and Schmidt]{ilharco2021openclip}
Gabriel Ilharco, Mitchell Wortsman, Ross Wightman, Cade Gordon, Nicholas Carlini, Rohan Taori, Achal Dave, Vaishaal Shankar, Hongseok Namkoong, John Miller, Hannaneh Hajishirzi, Ali Farhadi, and Ludwig Schmidt.
\newblock {OpenCLIP}, 2021.

\bibitem[Jaeger et~al.(2023)Jaeger, Chitta, and Geiger]{jaeger2023hidden}
Bernhard Jaeger, Kashyap Chitta, and Andreas Geiger.
\newblock Hidden biases of end-to-end driving models.
\newblock In \emph{ICCV}, 2023.

\bibitem[Jia et~al.(2023{\natexlab{a}})Jia, Gao, Chen, Yan, Liu, and Li]{jia2023driveadapter}
Xiaosong Jia, Yulu Gao, Li Chen, Junchi Yan, Patrick~Langechuan Liu, and Hongyang Li.
\newblock {DriveAdapter}: Breaking the coupling barrier of perception and planning in end-to-end autonomous driving.
\newblock In \emph{ICCV}, 2023{\natexlab{a}}.

\bibitem[Jia et~al.(2023{\natexlab{b}})Jia, Wu, Chen, Xie, He, Yan, and Li]{jia2023think}
Xiaosong Jia, Penghao Wu, Li Chen, Jiangwei Xie, Conghui He, Junchi Yan, and Hongyang Li.
\newblock Think twice before driving: Towards scalable decoders for end-to-end autonomous driving.
\newblock In \emph{CVPR}, 2023{\natexlab{b}}.

\bibitem[Jia et~al.(2024)Jia, Yang, Li, Zhang, and Yan]{jia2024bench}
Xiaosong Jia, Zhenjie Yang, Qifeng Li, Zhiyuan Zhang, and Junchi Yan.
\newblock Bench2drive: Towards multi-ability benchmarking of closed-loop end-to-end autonomous driving.
\newblock In \emph{NeurIPS Datasets and Benchmarks Track}, 2024.

\bibitem[Jiang et~al.(2023)Jiang, Chen, Xu, Liao, Chen, Zhou, Zhang, Liu, Huang, and Wang]{jiang2023vad}
Bo Jiang, Shaoyu Chen, Qing Xu, Bencheng Liao, Jiajie Chen, Helong Zhou, Qian Zhang, Wenyu Liu, Chang Huang, and Xinggang Wang.
\newblock {VAD}: Vectorized scene representation for efficient autonomous driving.
\newblock In \emph{ICCV}, 2023.

\bibitem[Kuhn et~al.(2023)Kuhn, Gal, and Farquhar]{kuhn2023semantic}
Lorenz Kuhn, Yarin Gal, and Sebastian Farquhar.
\newblock Semantic uncertainty: Linguistic invariances for uncertainty estimation in natural language generation.
\newblock \emph{arXiv preprint arXiv:2302.09664}, 2023.

\bibitem[Lakshminarayanan et~al.(2016)Lakshminarayanan, Pritzel, and Blundell]{lakshminarayanan2017deepensemble}
Balaji Lakshminarayanan, Alexander Pritzel, and Charles Blundell.
\newblock Simple and scalable predictive uncertainty estimation using deep ensembles.
\newblock \emph{arXiv preprint arXiv:1612.01474}, 2016.

\bibitem[Li et~al.(2023)Li, Li, Wang, Zeng, Xu, Cai, Chen, Yan, Xu, Xiong, et~al.]{li2023open}
Hongyang Li, Yang Li, Huijie Wang, Jia Zeng, Huilin Xu, Pinlong Cai, Li Chen, Junchi Yan, Feng Xu, Lu Xiong, et~al.
\newblock Open-sourced data ecosystem in autonomous driving: the present and future.
\newblock \emph{arXiv preprint arXiv:2312.03408}, 2023.

\bibitem[Li et~al.(2024{\natexlab{a}})Li, Tseng, Girard, and Kolmanovsky]{li2024perceptionuncertainties}
Xiao Li, H.~Eric Tseng, Anouck Girard, and Ilya Kolmanovsky.
\newblock Autonomous driving with perception uncertainties: Deep-ensemble based adaptive cruise control.
\newblock \emph{arXiv preprint arXiv:2403.15577}, 2024{\natexlab{a}}.

\bibitem[Li et~al.(2024{\natexlab{b}})Li, Li, Wang, Lan, Yu, Ji, Li, Zhu, Kautz, Wu, et~al.]{li2024hydra}
Zhenxin Li, Kailin Li, Shihao Wang, Shiyi Lan, Zhiding Yu, Yishen Ji, Zhiqi Li, Ziyue Zhu, Jan Kautz, Zuxuan Wu, et~al.
\newblock {Hydra-MDP}: End-to-end multimodal planning with multi-target hydra-distillation.
\newblock \emph{arXiv preprint arXiv:2406.06978}, 2024{\natexlab{b}}.

\bibitem[Li et~al.(2024{\natexlab{c}})Li, Yu, Lan, Li, Kautz, Lu, and Alvarez]{li2024ego}
Zhiqi Li, Zhiding Yu, Shiyi Lan, Jiahan Li, Jan Kautz, Tong Lu, and Jose~M Alvarez.
\newblock Is ego status all you need for open-loop end-to-end autonomous driving?
\newblock In \emph{CVPR}, 2024{\natexlab{c}}.

\bibitem[Liang et~al.(2023)Liang, He, and Tan]{Liang2023ACS}
Jian Liang, Ran He, and Tien-Ping Tan.
\newblock A comprehensive survey on test-time adaptation under distribution shifts.
\newblock \emph{arXiv preprint arXiv:2303.15361}, 2023.

\bibitem[Liang et~al.(2024)Liang, He, and Tan]{liang2024comprehensive}
Jian Liang, Ran He, and Tieniu Tan.
\newblock A comprehensive survey on test-time adaptation under distribution shifts.
\newblock \emph{IJCV}, 2024.

\bibitem[Liao et~al.(2025)Liao, Chen, Yin, Jiang, Wang, Yan, Zhang, Li, Zhang, Zhang, et~al.]{liao2025diffusiondrive}
Bencheng Liao, Shaoyu Chen, Haoran Yin, Bo Jiang, Cheng Wang, Sixu Yan, Xinbang Zhang, Xiangyu Li, Ying Zhang, Qian Zhang, et~al.
\newblock Diffusiondrive: Truncated diffusion model for end-to-end autonomous driving.
\newblock In \emph{CVPR}, 2025.

\bibitem[Liu and Feng(2022)]{Liu2022CurseOR}
Henry~X. Liu and Shuo Feng.
\newblock Curse of rarity for autonomous vehicles.
\newblock \emph{Nature Communications}, 15, 2022.

\bibitem[Loshchilov(2017)]{loshchilov2017decoupled}
I Loshchilov.
\newblock Decoupled weight decay regularization.
\newblock \emph{arXiv preprint arXiv:1711.05101}, 2017.

\bibitem[Marsden et~al.(2024)Marsden, D{\"o}bler, and Yang]{marsden2024universal}
Robert~A Marsden, Mario D{\"o}bler, and Bin Yang.
\newblock Universal test-time adaptation through weight ensembling, diversity weighting, and prior correction.
\newblock 2024.

\bibitem[Najm et~al.(2007)Najm, Smith, Yanagisawa, et~al.]{najm2007pre}
Wassim~G Najm, John~D Smith, Mikio Yanagisawa, et~al.
\newblock Pre-crash scenario typology for crash avoidance research.
\newblock Technical report, United States. Department of Transportation. NHTSA, 2007.

\bibitem[OpenAI(2024)]{openaio1}
OpenAI.
\newblock {OpenAI} o1 system card.
\newblock https://openai.com/index/openai-o1-system-card/, 2024.

\bibitem[Park et~al.(2021)Park, Ambrus, Guizilini, Li, and Gaidon]{park2021pseudo}
Dennis Park, Rares Ambrus, Vitor Guizilini, Jie Li, and Adrien Gaidon.
\newblock Is pseudo-lidar needed for monocular 3d object detection?
\newblock In \emph{ICCV}, 2021.

\bibitem[Park et~al.(2024)Park, Jeong, Yoon, Jeong, and Yoon]{park2024t4p}
Daehee Park, Jaeseok Jeong, Sung-Hoon Yoon, Jaewoo Jeong, and Kuk-Jin Yoon.
\newblock T4p: Test-time training of trajectory prediction via masked autoencoder and actor-specific token memory.
\newblock In \emph{CVPR}, 2024.

\bibitem[Prakash et~al.(2021)Prakash, Chitta, and Geiger]{Prakash2021transfuser}
Aditya Prakash, Kashyap Chitta, and Andreas Geiger.
\newblock Multi-modal fusion transformer for end-to-end autonomous driving.
\newblock In \emph{CVPR}, 2021.

\bibitem[Sadat et~al.(2020)Sadat, Casas, Ren, Wu, Dhawan, and Urtasun]{Sadat2020P3}
Abbas Sadat, Sergio Casas, Mengye Ren, Xinyu Wu, Pranaab Dhawan, and Raquel Urtasun.
\newblock Perceive, predict, and plan: Safe motion planning through interpretable semantic representations.
\newblock In \emph{ECCV}, 2020.

\bibitem[Shalev-Shwartz et~al.(2017)Shalev-Shwartz, Shammah, and Shashua]{shalev2017safe}
Shai Shalev-Shwartz, Shaked Shammah, and Amnon Shashua.
\newblock On a formal model of safe and scalable self-driving cars.
\newblock \emph{arXiv preprint arXiv:1708.06374}, 2017.

\bibitem[Shannon(1948)]{shannon1948entropy}
Claude~Elwood Shannon.
\newblock A mathematical theory of communication.
\newblock \emph{The Bell System Technical Journal}, pages 379--423, 1948.

\bibitem[Shu et~al.(2022)Shu, Nie, Huang, Yu, Goldstein, Anandkumar, and Xiao]{shu2022test}
Manli Shu, Weili Nie, De-An Huang, Zhiding Yu, Tom Goldstein, Anima Anandkumar, and Chaowei Xiao.
\newblock Test-time prompt tuning for zero-shot generalization in vision-language models.
\newblock In \emph{NeurIPS}, 2022.

\bibitem[Sima et~al.(2024)Sima, Renz, Chitta, Chen, Zhang, Xie, Luo, Geiger, and Li]{sima2023drivelm}
Chonghao Sima, Katrin Renz, Kashyap Chitta, Li Chen, Hanxue Zhang, Chengen Xie, Ping Luo, Andreas Geiger, and Hongyang Li.
\newblock {DriveLM}: Driving with graph visual question answering.
\newblock In \emph{ECCV}, 2024.

\bibitem[Sun et~al.(2024)Sun, Li, Dalal, Xu, Vikram, Zhang, Dubois, Chen, Wang, Koyejo, et~al.]{sun2024learning}
Yu Sun, Xinhao Li, Karan Dalal, Jiarui Xu, Arjun Vikram, Genghan Zhang, Yann Dubois, Xinlei Chen, Xiaolong Wang, Sanmi Koyejo, et~al.
\newblock Learning to (learn at test time): Rnns with expressive hidden states.
\newblock \emph{arXiv preprint arXiv:2407.04620}, 2024.

\bibitem[Thorn et~al.(2018)Thorn, Kimmel, Chaka, Hamilton, et~al.]{thorn2018framework}
Eric Thorn, Shawn~C Kimmel, Michelle Chaka, Booz~Allen Hamilton, et~al.
\newblock A framework for automated driving system testable cases and scenarios.
\newblock Technical report, United States. Department of Transportation. NHTSA, 2018.

\bibitem[Tong et~al.(2023)Tong, Sima, Wang, Chen, Wu, Deng, Gu, Lu, Luo, Lin, and Li]{tong2023occnet}
Wenwen Tong, Chonghao Sima, Tai Wang, Li Chen, Silei Wu, Hanming Deng, Yi Gu, Lewei Lu, Ping Luo, Dahua Lin, and Hongyang Li.
\newblock Scene as occupancy.
\newblock In \emph{ICCV}, 2023.

\bibitem[Vitelli et~al.(2022)Vitelli, Chang, Ye, Ferreira, Wołczyk, Osiński, Niendorf, Grimmett, Huang, Jain, and Ondruska]{vitelli2022safetynet}
Matt Vitelli, Yan Chang, Yawei Ye, Ana Ferreira, Maciej Wołczyk, Błażej Osiński, Moritz Niendorf, Hugo Grimmett, Qiangui Huang, Ashesh Jain, and Peter Ondruska.
\newblock {SafetyNet}: Safe planning for real-world self-driving vehicles using machine-learned policies.
\newblock In \emph{ICRA}, 2022.

\bibitem[Wang et~al.(2021)Wang, Shelhamer, Liu, Olshausen, and Darrell]{wang2020tent}
Dequan Wang, Evan Shelhamer, Shaoteng Liu, Bruno Olshausen, and Trevor Darrell.
\newblock Tent: Fully test-time adaptation by entropy minimization.
\newblock In \emph{ICLR}, 2021.

\bibitem[Wang et~al.(2022)Wang, Fink, Van~Gool, and Dai]{wang2022continual}
Qin Wang, Olga Fink, Luc Van~Gool, and Dengxin Dai.
\newblock Continual test-time domain adaptation.
\newblock In \emph{CVPR}, 2022.

\bibitem[Wang et~al.(2023)Wang, Zhu, Huang, Chen, Zhu, and Lu]{wang2023drivedreamer}
Xiaofeng Wang, Zheng Zhu, Guan Huang, Xinze Chen, Jiagang Zhu, and Jiwen Lu.
\newblock {DriveDreamer}: Towards real-world-driven world models for autonomous driving.
\newblock \emph{arXiv preprint arXiv:2309.09777}, 2023.

\bibitem[Wayve(2024)]{waywe2024ad2}
Wayve.
\newblock Wayve's av2.0 approach: Pioneering the end-to-end ai driving model.
\newblock \url{https://youtu.be/AEfq5nFi7s8}, 2024.

\bibitem[Weng et~al.(2024)Weng, Ivanovic, Wang, Wang, and Pavone]{Weng2024para}
Xinshuo Weng, Boris Ivanovic, Yan Wang, Yue Wang, and Marco Pavone.
\newblock {PARA-Drive}: Parallelized architecture for real-time autonomous driving.
\newblock In \emph{CVPR}, 2024.

\bibitem[Wu et~al.(2022)Wu, Jia, Chen, Yan, Li, and Qiao]{wu2022tcp}
Penghao Wu, Xiaosong Jia, Li Chen, Junchi Yan, Hongyang Li, and Yu Qiao.
\newblock Trajectory-guided control prediction for end-to-end autonomous driving: A simple yet strong baseline.
\newblock In \emph{NeurIPS}, 2022.

\bibitem[Yang et~al.(2024)Yang, Gao, Qiu, Chen, Li, Dai, Chitta, Wu, Zeng, Luo, et~al.]{yang2024generalized}
Jiazhi Yang, Shenyuan Gao, Yihang Qiu, Li Chen, Tianyu Li, Bo Dai, Kashyap Chitta, Penghao Wu, Jia Zeng, Ping Luo, et~al.
\newblock Generalized predictive model for autonomous driving.
\newblock In \emph{CVPR}, 2024.

\bibitem[Yoon et~al.(2024)Yoon, Yoon, Tee, Hasegawa-Johnson, Li, and Yoo]{yoon2024c}
Hee~Suk Yoon, Eunseop Yoon, Joshua Tian~Jin Tee, Mark Hasegawa-Johnson, Yingzhen Li, and Chang~D Yoo.
\newblock C-tpt: Calibrated test-time prompt tuning for vision-language models via text feature dispersion.
\newblock In \emph{ICLR}, 2024.

\bibitem[Zimmerlin et~al.(2024)Zimmerlin, Beißwenger, Jaeger, Geiger, and Chitta]{zimmerlin2024tf}
Julian Zimmerlin, Jens Beißwenger, Bernhard Jaeger, Andreas Geiger, and Kashyap Chitta.
\newblock Hidden biases of end-to-end driving datasets.
\newblock \emph{arXiv preprint arXiv:2412.09602}, 2024.

\end{thebibliography}
}

\clearpage
\setcounter{page}{1}
\maketitlesupplementary
\renewcommand\thesection{\Alph{section}}
\setcounter{section}{0}

    In this supplementary document, we present additional related work, methodological details, experimental details, as well as more experimental results and visualizations. 

\section{Additional Related Work}

Our work extends Hydra-MDP~\cite{li2024hydra}, a state-of-the-art trajectory-scoring planner. Building upon the multimodal perception architecture TransFuser~\cite{chitta2023transfuser}, Hydra-MDP learns from multiple teachers, including the rule-based PDM planner~\cite{dauner2023parting}, through a knowledge distillation framework. 
Besides some ensembling techniques~\cite{lakshminarayanan2017deepensemble,li2024perceptionuncertainties}, very few works in this domain explore the potential of adding more test-time compute to improve the performance.
In our work, Centaur enables refined planning via test-time training.

\boldparagraph{Semantic Entropy} One of our baselines, the semantic entropy, is inspired by semantic uncertainty~\cite{kuhn2023semantic} in LLMs, where uncertainty estimation in language models is improved by addressing the limitations of traditional entropy measures using clustering to reduce redundant uncertainty with linguistic invariance. Semantic entropy is calculated based on probabilities across clusters rather than individual responses. This captures the uncertainty across genuinely different meanings rather than superficial wording differences. In our case, we adopt a similar strategy to cluster and formulate entropy over the predicted trajectories.

\boldparagraph{Safety-Critical Scenarios} Evaluating end-to-end autonomous driving models presents significant challenges. While traditional metrics like average displacement error (ADE) and final displacement error (FDE) have been widely used~\cite{hu2023uniad, jiang2023vad,hu2023gaia}, recent research~\cite{li2024ego} has highlighted their limitations in comprehensively assessing autonomous driving systems. This has led to the adoption of more sophisticated benchmarks like NAVSIM~\cite{dauner2024navsim}, which evaluates models across multiple dimensions, including safety, progress, and traffic rule compliance. Bench2Drive~\cite{jia2024bench} is a newly-proposed benchmark designed to evaluate the diverse capabilities of end-to-end autonomous driving systems in a closed-loop setting, which tests driving models to handle 44 different individual scenarios. InterPlan~\cite{hallgarten2024interplan} constructs out-of-distribution testing scenarios like dealing with a crush site in the middle of the road, and proposes using large language models to adapt parameters of non-differentiable planners at test time. In our work, \texttt{navsafe} is built upon \texttt{navtest}, delving deeper into rare cases that the state-of-the-art planners may fail to show the performance gap in detail. Unlike InterPlan, we focus on end-to-end driving. Unlike Bench2Drive, \texttt{navsafe} is based on real-world data.

\boldparagraph{Evidential Uncertainty} Amini et al.~\cite{amini2020deepevidential} propose Deep Evidential Regression for training deterministic neural networks to estimate both continuous targets and associated uncertainties. This approach enables the network to infer the hyperparameters of evidential distributions, thereby capturing both aleatoric (data) and epistemic (model) uncertainties without requiring sampling during inference or out-of-distribution (OOD) examples for training. By applying regularization to ensure alignment between predicted evidence and actual outputs, their method produces well-calibrated uncertainty estimates. It is scalable and robust, demonstrating resistance to adversarial attacks and OOD samples, making it suitable for complex tasks where uncertainty estimation is crucial. In this supplementary document, we adapt the evidential regression loss for  TransFuser~\cite{chitta2023transfuser}, to evaluate its suitability for TTT.

\begin{table}[t!]
  \centering
  \begin{tabular}{|c|c|}
    \hline
    Notation & Description \\
    \hline
    $\mathcal{L}_{im}$ & Imitation loss  \\
    $\mathcal{L}_{kd}$ & Knowledge distillation  loss \\
    $\bt$ & Trajectory \\
    $\bp$ & Waypoint in trajectory \\
    $\pi$ & Policy for planning \\
    $\bx$ & Sensor inputs \\
    $\bc$ & Navigation command \\
    $\theta$ & Model parameters \\
    $H$ & Cluster entropy \\
    $M$ & Number of trajectory candidates \\
    $score_a$ & Anchor score per cluster \\
    $k$ & Planning vocabulary size \\
    $\gamma, \alpha, \beta, \upsilon$ & Evidential parameters \\
    $F$ & Number of history frames \\
    $\eta$ & Learning rate \\
    $\tau$ & Clustering threshold \\
    \hline
  \end{tabular}
  \caption{The notation used in our paper, with descriptions.}
  \label{tab:notation}
\end{table}

\section{Additional Methodological Details}

In this section, we present an adaptation of Semantic Entropy for enabling TTT with trajectory regression based approaches. We then provide additional details about the loss function for trajectory scoring, and the modeling of Semantic Entropy for both trajectory scoring and trajectory regression methods. We provide a table of all notation in \tabref{tab:notation}.

\subsection{Trajectory Regression Planner}

In order to enable uncertainty estimation in the trajectory regression setting, we predict the `evidential parameters'~\cite{amini2020deepevidential} of a Normal Inverse Gamma (NIG) distribution $\alpha$, $\beta$, $\gamma$, and $\upsilon$ for each scalar dimension $y_i$ of the output using $\pi_\theta$, instead of a standard point estimate. We can sample from this distribution as:
\begin{align}
    (y_1, \dots, y_N) &\sim \mathcal{N}(\mu, \sigma^2)\nonumber,\\ 
    \mu \sim \mathcal{N}(\gamma, \sigma^2 \upsilon^{-1}), 
    \quad & \qquad
    \sigma^2 \sim \Gamma^{-1}(\alpha, \beta),
    \label{eqn:evidential}
\end{align}
where $\Gamma(\cdot)$ is the gamma function to implement an Inverse-Gamma prior and $\mathcal{N}(\cdot)$ is the Gaussian distribution. The model's output trajectory for the planning task is taken as the mean value $\gamma$ of the NIG distribution, resulting in a single trajectory $\hat{{\bf {t}}}_0 = \pi_\theta({\bf {x}}_0) = (\gamma_0, \gamma_1, \cdots , \gamma_N)$. Learning the policy in this setting is referred to as imitation learning: 
\begin{equation}
    \label{eqn:objective}
   \arg \min_{\theta} \mathbb{E}_{(\bx,\bt,\bc)\sim D_{im}} [\mathcal{L}_{im}(\bt, \pi_\theta(\bx,\bc))],
\end{equation}
where $D_{im} = \{(\bx,\bt,\bc), \cdots \}$ is a dataset collected by an expert driver controlling the vehicle while interacting with the other traffic participants. We use the evidential loss for $\mathcal{L}_{im}$, which is a maximum likelihood objective for the evidential parameters~\cite{amini2020deepevidential}.

As the required output of a trajectory regression planner is a vector, which we denote $\vec{\bt} = (\bp_0, \bp_1, \cdots , \bp_N)$, our `evidential parameters'~\cite{amini2020deepevidential} are in fact in a vectorized format.
We represent the evidential parameters of a Normal Inverse Gamma (NIG) distribution $\vec{\alpha}$, $\vec{\beta}$, $\vec{\gamma}$, and $\vec{\upsilon}$ for the original output $\vec{\bt}$ using our policy $\pi_\theta$. We can sample from this distribution as:
\begin{align}
    (\vec{\bt}_1, \dots, \vec{\bt}_N) &\sim \mathcal{N}(\vec{\mu}, \vec{\sigma^2})\nonumber,\\ 
    \vec{\mu} \sim \mathcal{N}(\vec{\gamma}, \vec{\sigma^2 \upsilon^{-1}} ), 
    \quad & \qquad
    \vec{\sigma^2} \sim \Gamma^{-1}(\vec{\alpha}, \vec{\beta}),
    \label{eqn:evidential-vector}
\end{align}
where $\Gamma(\cdot)$ is the gamma function to implement an Inverse-Gamma prior and $\mathcal{N}(\cdot)$ is the Gaussian distribution. The model's output trajectory for the planning task is taken as the mean value $\vec{\gamma}$ of the NIG distribution, resulting in a single trajectory $\hat{{\bf {t}}}_0 = \pi_\theta({\bf {x}}_0) = \vec{\gamma}$. Learning the policy in this setting is referred to as imitation learning: 
\begin{equation}
    \label{eqn:objective-imi-kd}
   \arg \min_{\theta} \mathbb{E}_{(\bx,\vec{\bt},\bc)\sim D_{im}} [\mathcal{L}_{im}(\vec{\bt}, \pi_\theta(\bx,\bc))],
\end{equation}
where $D_{im} = \{(\bx,\vec{\bt},\bc), \cdots \}$ is a dataset collected by an expert driver controlling the vehicle while interacting with the other traffic participants.
We use the evidential loss for $\mathcal{L}_{im}$, which is a maximum likelihood objective for the evidential parameters~\cite{amini2020deepevidential}.

\subsection{Trajectory Scoring Planner}
In our implementation of Hydra-MDP~\cite{li2024hydra}, aside from the cross-entropy loss $\mathcal{L}_{kd}$ in Eq.~\eqref{eqn:objective-score}, we include an imitation loss $\mathcal{L}_{im}$ to regularize the trajectory distribution, following VADv2~\cite{chen2024vadv2}. The overall loss function is:
\begin{equation}
\label{eqn:multi-score}
    \mathcal{L}=\sum_j\mathcal{L}_{im}(\Tilde{\bt}, \pi_\theta(\bt_j,\bx,\bc)) + \mathcal{L}_{kd}(e(\bt,\bx,\bc), \pi_\theta(\bt,\bx,\bc))
\end{equation}
where the imitation loss ($\mathcal{L}_{im}$) is implemented as a distance-based cross-entropy loss~\cite{li2024hydra}. The query for each trajectory is supervised by the imitation target produced by the L2 distance between that trajectory and the one in the log-replay. The softmax function is applied on such a distance distribution to produce a probability distribution.

\begin{table*}[t!]
    \centering
    \caption{\textbf{NAVSIM v1.0 comparison to state-of-the-art.} Unlike the official leaderboard from \tabref{tab:navtest_all} in the main paper, here we report results for models with a single training seed on the earlier NAVSIM version used for the 2024 NAVSIM Challenge. Transfuser-SE denotes equipping Transfuser with TTT and semantic entropy.}
    \label{tab:navtest_all_supp}
    \small %
    \begin{tabular}{ll|cccc|cc}
    \toprule
    \textbf{Model} & \textbf{Deployment} & \textbf{NC} $\uparrow$ & \textbf{DAC} $\uparrow$ & \textbf{EP} $\uparrow$ & \textbf{C} $\uparrow$ & \textbf{TTC} $\uparrow$ & \textbf{PDMS} $\uparrow$ \\
    \midrule
    \multirow{2}{*}{UniAD~\cite{hu2023uniad}} & None & $97.8$ & $91.9$ & $78.8$ & $100.0$ & $92.9$ & $83.4$ \\
    & Optimization & $95.4$ & $86.0$ & $71.9$ & $99.3$ & $87.3$ & $75.2$ \\
    \midrule
    PARA-Drive~\cite{Weng2024para} & None & $97.9$ & $92.4$ & $79.3$ & $99.8$ & $93.0$ & $84.0$  \\
    + Occupancy Head & None & $97.8$ & $92.0$ & $78.3$ & $99.7$ & $93.7$ & $83.6$ \\
    \midrule
    TransFuser~\cite{chitta2023transfuser} & None & $97.1$ & $91.9$ & $79.3$ & $100.0$ & $91.9$ & $83.4$ \\
    TransFuser-SE & TTT & $97.9$ & $92.9$ & $80.1$ & $100.0$ & $94.1$ & $85.7$ \\

    \midrule
    DiffusionDrive~\cite{liao2025diffusiondrive} & None & $98.2$ & $96.2$ & $82.2$ & $100.0$ & $94.7$ & $ 88.1$ \\
    \midrule
    Hydra-MDP~\cite{li2024hydra} & None & $98.4$ & $97.8$ & $86.5$ & $100.0$ & $93.9$ & $90.3$ \\
    Hydra-SE & TTT & $99.2$ & $98.5$ & $85.7$ & $100.0$ & $97.1$ & $91.8$ \\
    Centaur & TTT & $99.5$ & $98.9$ & $85.9$ & $100.0$ & $98.0$ & $92.6$ \\
    \bottomrule
    \end{tabular}
\end{table*}

\subsection{Entropy Estimation}

\noindent\textbf{Sampling of anchors and candidates.} We first sample $M=100$ trajectory candidates from the planning vocabulary with a size of 8192. The sampling process is a weighted random sampling where the weight is the average ground truth PDMS of each trajectory on the \texttt{navtrain} split. Note that this is similar to the weight calculation we use in our fallback layer baseline. We then sample 5 anchors from the trajectory candidates via domain-specific criteria, following DriveLM~\cite{sima2023drivelm}. Unlike standard clustering algorithms like k-means, our approach is computationally efficient and differentiable. Further, due to the choice of fixed anchors, it ensures coverage of a large amount of drivable area, independent of the choice of feature space. 

\boldparagraph{Semantic Entropy} Intuitively, when a policy expects many trajectories to result in similar outcomes, it is more `certain' than when it expects most trajectories to result in diverse outcomes. Semantic Entropy computes the nearest anchor to each of our $M$ trajectory candidates (in terms of $L_2$ distance) in a feature space, rather than the Euclidean space of trajectories. While the choice of feature space to use with our approach is flexible, in this adaptation, we prioritize simplicity. For our trajectory scoring planner, we directly use their output score features as the choice of feature space. For trajectory regression, we consider two simple score features, $\{ L_{\gamma}, L_{\mu} \}$, based on the evidential parameters $\alpha$, $\beta$, $\gamma$, and $\upsilon$. We first sample $N$ values for $\sigma$ via $\Gamma^{-1}(\alpha, \beta)$ and sample the corresponding $\mu$ via $\mathcal{N}(\gamma, \sigma^2 \upsilon^{-1})$. Following this, for each trajectory in the $M$ candidates, we calculate its L2 distance $L_{\gamma}$ to $\gamma$ and L2 distance $L_{\mu}$ to the closest $\mu$ of the $N$ values. After identifying the nearest anchor for all $M$ candidates in the selected feature space, we define \uncname (denoted as $H$) to be the Shannon Entropy of the resulting 5-way categorical distribution.

\noindent\textbf{Clustering Threshold for Semantic Entropy.}
Here we further detail the process of clustering in Semantic Entropy, where we use $\tau$ as a threshold.
Clustering of the trajectory candidates to the trajectory anchors requires a distance function, $D(\cdot, \cdot)$, to induce the equivalence between two trajectories in terms of score features. 
We adopt $L_2$ distance as our implementation of $D$ for its computational efficiency and compatibility to the score features from both types of planners.
The clustering is a fixed-center one-step k-means. we use $D$ to measure the score features of two trajectories (one from the anchor, the other from the candidate), and if it is under a certain threshold $\tau$ then they go into one cluster. Trajectories are assigned to the closest center which is measured via $D$ over the score features.
The reason of using fixed-starting centers in the clustering is that they will cover enough drivable area whatever the score features are.
In practice, we set $\tau$ to be sufficiently large, minimizing the number of trajectories that fall outside the threshold.  
For these remaining trajectories, we assign a default cluster, based on the closest anchor in terms of $L_2$ distance in trajectory space.
\begin{table*}[t!]
    \centering
    \caption{\textbf{Ablations on \texttt{navtest}.} More results, in addition to  \tabref{tab:navtest_ablation} in the main paper. TTT$^{*}$ uses the SVD of history gradients to combine them, instead of averaging. Hydra-SE$^\textbf{f}$ refers to using $f$ history frames. \colorbox{gray!50}{Hydra-SE$^\textbf{4}$} denotes TTT being applied only on inference of selected frames where uncertainty is higher than a threshold (\textit{e.g.} \textbf{Main} L383). We report \textbf{amortized latency} on \texttt{navtest}. }
    \label{tab:navtest_ablation_supp}
    \small %
    \setlength{\tabcolsep}{2mm}
    \begin{tabular}{l|ll|rrrr|rr|c}
    \toprule
    \textbf{Model} & \textbf{Deployment} & \textbf{Uncertainty} & \textbf{NC} $\uparrow$ & \textbf{DAC} $\uparrow$ & \textbf{EP} $\uparrow$ & \textbf{C} $\uparrow$ & \textbf{TTC} $\uparrow$ & \textbf{PDMS} $\uparrow$ & \textbf{Latency (ms)} $\downarrow$ \\
    \midrule
    \multirow{4}{*}{TransFuser~\cite{chitta2023transfuser}} & - & - & $97.1$ & $91.9$ & $79.3$ & $100.0$ & $91.9$ & $83.4$ & $103.1$ \\ 
    & Fallback Layer & Evidential & $98.8$ & $97.4$ & $13.1$ & $100.0$ & $97.9$ & $60.6$ & $108.7$ \\
    & Fallback Layer & Semantic Entropy & $99.2$ & $97.2$ & $14.6$ & $99.8$ & $96.5$ & $59.9$ & $109.9$ \\
    & TTT & Evidential & $97.9$ & $92.5$ & $79.6$ & $100.0$ & $92.9$ & $84.2$ & $149.3$ \\
    \midrule
    \multirow{2}{*}{TransFuser-SE} & TTT & \multirow{2}{*}{Semantic Entropy} & $97.9$ & $92.9$ & $80.1$ & $100.0$ & $94.1$ & $85.7$ & $153.8$ \\
    & TTT$^{*}$ & & $97.4$ & $89.3$ & $78.3$ & $100.0$ & $91.1$ & $83.8$ & $157.6$ \\
    \midrule
    \multirow{4}{*}{Hydra-MDP~\cite{li2024hydra}} & - & - & $98.4$ & $97.8$ & $86.5$ & $100.0$ & $93.9$ & $90.3$ & $243.9$ \\
    & Fallback Layer & KL Divergence & $99.1$ & $99.2$ & $9.6$ & $100.0$ & $96.7$ & $63.1$ & $263.2$ \\
    & Fallback Layer & Semantic Entropy & $99.7$ & $99.5$ & $10.4$ & $100.0$ & $98.9$ & $65.3$ & $256.4$ \\
    & TTT & KL Divergence & $98.9$ & $98.1$ & $86.0$ & $100.0$ & $94.4$ & $91.5$ & $357.1$ \\
    \midrule
    \multirow{4}{*}{Hydra-MDP-ViT-L} & - & - & $98.4$ & $97.7$ & $85.0$ & $100.0$ & $94.5$ & $89.9$ & $237.2$ \\
    & Fallback Layer & KL Divergence & $99.7$ & $99.2$ & $10.4$ & $100.0$ & $98.9$ & $67.4$ & $259.5$ \\
    & Fallback Layer & Semantic Entropy & $99.6$ & $97.5$ & $11.8$ & $100.0$ & $94.1$ & $61.0$ & $255.7$ \\
    & TTT & KL Divergence & $99.1$ & $98.2$ & $89.6$ & $100.0$ & $96.7$ & $90.9$ & $354.1$ \\
    \midrule
    Hydra-SE$^\textbf{4}$  & \multirow{4}{*}{TTT} & \multirow{4}{*}{Semantic Entropy} & $99.2$ & $98.5$ & $85.7$ & $100.0$ & $97.1$ & $91.8$ & \multirow{4}{*}{$312.5$} \\
    Hydra-SE$^\textbf{3}$ & & & $98.9$ & $98.7$ & $86.7$ & $100.0$ & $95.1$ & $91.5$ & \\
    Hydra-SE$^\textbf{2}$ & & & $98.8$ & $98.6$ & $86.5$ & $100.0$ & $95.0$ & $91.3$ & \\
    Hydra-SE$^\textbf{1}$ & & & $98.9$ & $98.6$ & $86.7$ & $100.0$ & $95.1$ & $91.4$ & \\
    \midrule
    \rowcolor{gray!50} Hydra-SE$^\textbf{4}$ & TTT & Semantic Entropy & $98.6$ & $98.1$ & $86.3$ & $100.0$ & $94.9$ & $91.3$ & ${279.6}$ \\
    \bottomrule
    \end{tabular}
\end{table*}

\section{Experimental Details}

\boldparagraph{Hyper-parameters} For all the experiments with Hydra-MDP, we adopt 8192 as the size of planning vocabulary, following the state-of-the-art setting~\cite{li2024hydra}. For all the experiments with TransFuser, we equip each waypoint regression head with a evidential regression loss, following the implementation in Evidential Deep Learning~\cite{amini2020deepevidential}. The output for both trajectory scoring and trajectory regression planners is a 40-waypoint trajectory over 4 seconds, sampled at 10 Hz, with each waypoint defined by their x, y, and heading values. $\tau=0.06$ for Hydra-SE, $\tau=0.02$ for TransFuser-SE. 
For Transfuser, we use 1 $\times$ NVIDIA A100, batch size as 64, LR as 1e-4.

\subsection{Motivation of Baselines}
\begin{figure*}[t!]
    \centering
    \includegraphics[width=0.8\linewidth]{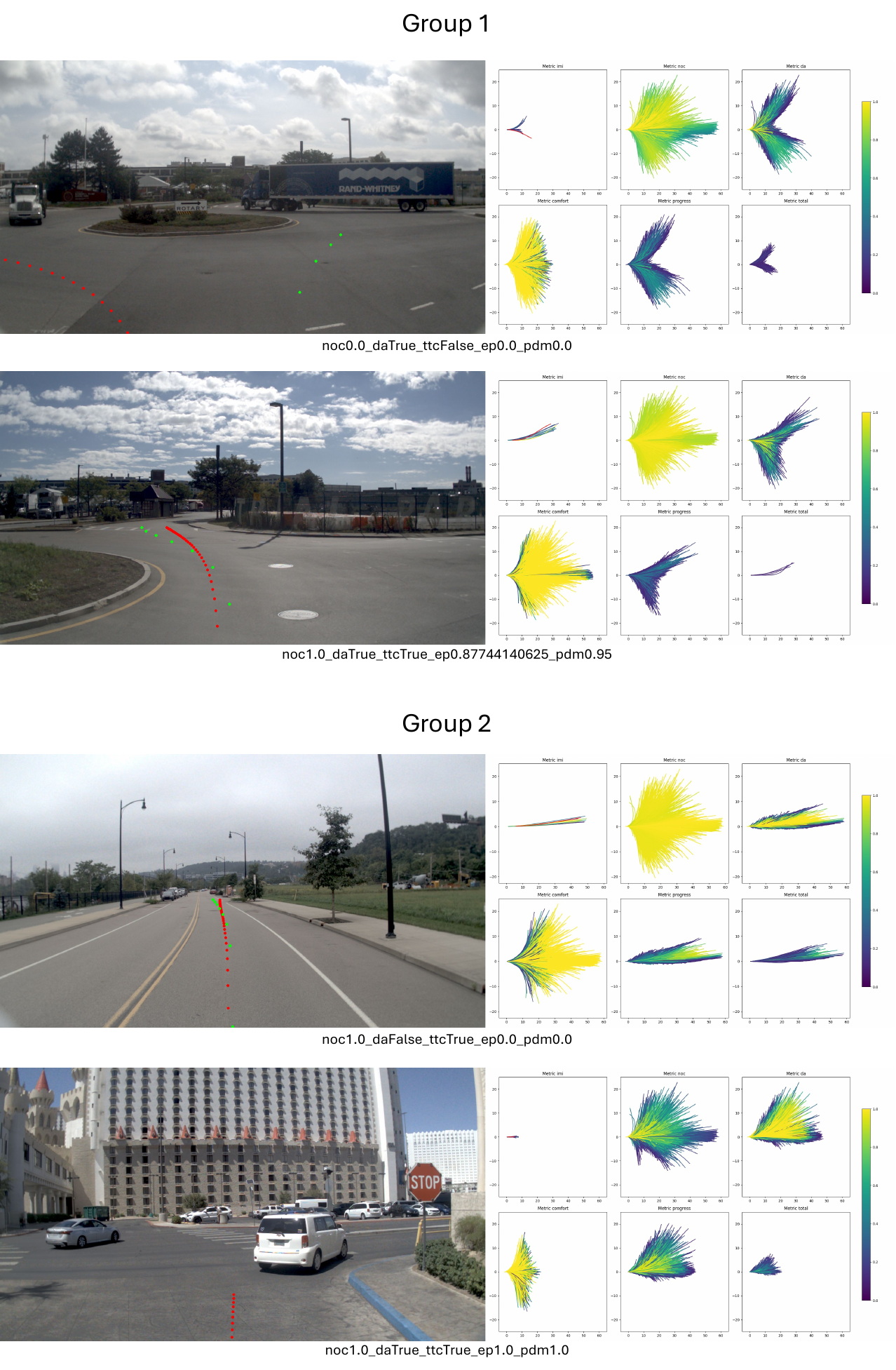}
    \caption{\textbf{Score distribution differences between success and failure cases.} We show 2 groups of comparisons with similar scenario types within each group. We observe larger diversities among the groups in failure cases (upper row in group) compared to successes.}
    \label{fig:vis_group}
\end{figure*}

We select the baselines in uncertainty estimation and deployment strategy based on the following motivations:
\begin{itemize}
    \item \textbf{KL Divergence uncertainty as baseline for uncertainty in trajectory scoring planners:} this is motivated via a visualization of the score feature distributions in the success and failure cases. We found inconsistencies between the score distributions from different metrics, especially the scores from NC and DAC, as shown in \figref{fig:vis_group}. Thus, we wonder if such an inconsistency can serve as an indicator for possible failure cases.
    \item \textbf{Evidential uncertainty as baseline for uncertainty in trajectory regression planner:} as many regression problems use evidential uncertainty as a baseline method, we follow this default setting as a baseline for TTT with TransFuser.
    \item \textbf{Fallback layer as baseline for deployment strategy in end-to-end planner:} We adopt a fallback layer in comparison to test-time training as it is a common practice in industry, and unlike test-time trajectory optimization, it can be implemented without access to explicit representations. The inference scheme used when deploying a fallback layer as the strategy is:
\begin{itemize}
    \item Infer all frames with the base end-to-end planners.
    \item Run uncertainty estimation for failure case classification over all frames to select the possible failure cases. Details of the classification setting are later discussed in \tabref{tab:fail_class}.
    \item For the frames selected as likely failures, run the fallback layer.
\end{itemize}
\end{itemize}

\subsection{Additional Ablation Studies}

\boldparagraph{NAVSIM 1.0} For completeness, we compare Centaur to additional baselines from NAVSIM 1.0 in \tabref{tab:navtest_all_supp}. We outperform all other methods. Here, we also include the results of TransFuser-SE, our baseline for the application of TTT to trajectory regression models. Similar to Centaur, it obtains an improvement of 2.3 PDMS relative to the base model TransFuser without TTT. Interestingly, UniAD~\cite{hu2023uniad}, which uses optimization with an expert-designed cost function, shows a deterioration in performance with this optimization step. This is due to a reduction in ego progress EP, much like our findings with a fallback layer in \tabref{tab:navtest_ablation}. Furthermore, the results of PARA-Drive~\cite{Weng2024para} show that incorporating an explicit representation of 3D occupancy from the end-to-end model can reduce its planning performance. These results show the need for an alternative means of ensuring safety during deployment, such as our proposed TTT.

\boldparagraph{Latency} We count the forward pass time in milliseconds, while for the planners with uncertainty estimation as well as a deployment strategy, we count the forward pass time once and then the corresponding time for the deployment method (test-time training or fallback layer). For test-time training we include the time for gradient calculation as well as the model update, and for the fallback layer we include the nearest neighbor selection for selecting the closest safe trajectory to the predicted trajectory via L2 distance. The hardware details for this inference test are as follows:
\begin{itemize}
    \item GPU: 1 $\times$ NVIDIA A100, driver version 535.54.03
    \item CPU: AMD EPYC 9124 16-Core Processor
    \item RAM: 500GB
\end{itemize}

As shown in \tabref{tab:navtest_ablation_supp}, a direct comparison of the base TransFuser and Hydra-MDP models, without integrating our proposed deployment and uncertainty enhancements, reveals a significant performance disparity in favor of Hydra-MDP. This can be attributed to their differing backbone architectures: TransFuser uses an efficient ResNet-34 backbone, which, although sacrificing performance, has a substantially lower latency of 103.1 ms compared to Hydra-MDP's 243.9 ms. In contrast, Hydra-MDP's more computationally expensive V2-99 backbone contributes to its higher performance by around 7 points in terms of PDMS. We re-iterate that our gradient calculation time is currently included in our latency measurement for TTT methods. 
It could be reduced by parallelizing execution in our implementation, or applying TTT on only selected frames where the uncertainty exceeds a threshold (\tabref{tab:navtest_ablation_supp}, bottom row).

\boldparagraph{TTT with SVD} 
Following~\cite{Gao2024fstta}, there is an alternative implementation for test-time adaptation that applies singular-value decomposition (SVD) to the history gradients to find the most useful ones for updating the model. As shown in \tabref{tab:navtest_ablation_supp}, TTT* (using SVD) has a lower performance on TransFuser-SE than TTT implemented with averaging gradients, in our case.

\boldparagraph{ViT-L Backbone} We ablate the influence of different backbones on Hydra-MDP under the settings of different uncertainty as well as deployment strategies in \tabref{tab:navtest_ablation_supp}. Overall the V2-99 backbone implemented in Hydra-MDP has better performance over the ViT-L backbone. However, on the fallback layer with KL Divergence setting, the ViT-L backbone surpasses V2-99 by around 4 points in terms of PDMS.

\boldparagraph{Buffer Length $F$} We show the influence of using different lengths of history frames in Hydra-SE on \texttt{navtest} in \tabref{tab:navtest_ablation_supp}. We observe a slight performance gain when increasing the length of history frames ($F$ as 2 has lowest performance). Overall, the performance with all settings of $F$ are consistent, indicating the robustness of our approach.

\begin{table}[t!]
    \centering
    \caption{\textbf{Failure identification.} More results in addition to  \tabref{tab:fail_class} in the main paper, with other planners and uncertainty thresholds.}
    \label{tab:fail_class_supp}
    \small %
    \setlength{\tabcolsep}{1.3mm}
    \begin{tabular}{l|l|c|rr}
    \toprule
    \textbf{Uncertainty} & \textbf{Model} & \textbf{Thresh.} & \textbf{TPR} $\uparrow$ & \textbf{Acc.} $\uparrow$
    \\
    \midrule
    \multirow{5}{*}{Evidential} & \multirow{5}{*}{TransFuser~\cite{chitta2023transfuser}} & $0.1$ & $47.2$ & $34.9$ \\
     &  & $0.3$ & $32.6$ & $48.2$ \\
     &  & $0.5$ & $26.9$ & $59.5$ \\
     &  & $0.7$ & $23.5$ & $62.8$ \\
     &  & $0.9$ & $7.4$ & $71.2$ \\
    \midrule
    KL Divergence & Hydra-MDP~\cite{li2024hydra}  & $1500$ & $41.3$ & $68.3$ \\
    \midrule
    \multirow{6}{*}{Semantic Entropy} & TransFuser~\cite{chitta2023transfuser} & $0.8$ & $35.3$ & $63.1$ \\
    \cmidrule{2-5}
    & \multirow{5}{*}{Hydra-MDP~\cite{li2024hydra}} & $0.2$ & $75.2$ & $61.9$ \\
    & & $0.5$ & $73.3$ & $62.4$ \\
    & & $0.8$ & ${62.5}$ & ${72.2}$ \\
    & & $1.1$ & $41.7$ & $73.2$\\
    & & $1.4$ & $10.9$ & $82.5$ \\
    \midrule
    \textit{Select All} & \multirow{2}{*}{\textit{TransFuser}~\cite{chitta2023transfuser}} & - & $\mathit{100.0}$ & $\mathit{24.1}$ \\
    \textit{Select None} &  & - & $\mathit{0.0}$ & $\mathit{75.9}$ \\
    \midrule
    \textit{Select All} & \multirow{2}{*}{\textit{Hydra-MDP}~\cite{li2024hydra}} & - & $\mathit{100.0}$ & $\mathit{15.6}$ \\
    \textit{Select None} &  & - & $\mathit{0.0}$ & $\mathit{84.4}$ \\
    \bottomrule
    \end{tabular}
\end{table}

\boldparagraph{Failure Identification Thresholds} We investigate the effect of different thresholds in the uncertainty identification for possible failure case identification. For both TransFuser with evidential uncertainty and Hydra-MDP with semantic entropy, we find that increasing the threshold results in a decrease in TPR for failure cases and an increase in overall accuracy. The reason is that on \texttt{navtest}, we have a long-tailed distribution that failure cases count for a small portion of the test samples, making it hard to pick true failure cases while not misidentifying true success cases. %

\end{document}